# A recurrent connectionist model of melody perception : An exploration using TRACX2

Daniel Defays[1], Robert M. French[2], Barbara Tillmann[2,3,4]

[1]Université of Liège, Department of Psychology, ddefays@uliege.be
[2]LEAD-CNRS UMR5022, French National Scientific Research Center (CNRS), robert.french@u-bourgogne.fr
[3]CNRS, UMR5292, INSERM, U1028, Lyon Neuroscience Research Center, Auditory Cognition and Psychoacoustics Team, Bron, France, barbara.tillmann@cnrs.fr
[4]University Lyon 1, Villeurbanne, France

Corresponding author
Robert M. French (robert.french@u-bourgogne.fr)
LEAD-CNRS UMR5022
French National Scientific Research Center (CNRS)
University of Burgundy,
21000 Dijon
France

## Abstract

Are similar, or even identical, mechanisms used in the computational modeling of speech segmentation, serial image processing and music processing? We address this question by exploring how TRACX2, (French et al., 2011; French & Cottrell, 2014; Mareschal & French, 2017), a recognition-based, recursive connectionist autoencoder model of chunking and sequence segmentation, which has successfully simulated speech and serial-image processing, might be applied to elementary melody perception. The model, a three-layer autoencoder that recognizes "chunks" of short sequences of intervals that have been frequently encountered on input, is trained on the tone intervals of melodically simple French children's songs. It dynamically incorporates the internal representations of these chunks into new input. Its internal representations cluster in a manner that is consistent with "human-recognizable" melodic categories. TRACX2 is sensitive to both contour and proximity information in the musical chunks that it encounters in its input. It shows the "end-of-word" superiority effect demonstrated by Saffran et al. (1999) for short musical phrases. The overall findings suggest that the recursive autoassociative chunking mechanism, as implemented in TRACX2, may be a general segmentation and chunking mechanism, underlying not only word- and image-chunking, but also elementary melody processing.

# 1. Introduction

Are similar, or even identical, mechanisms used in the computational modeling of speech segmentation, serial image processing and music processing? We address this question by exploring how TRACX2, (French et al., 2011; French & Cottrell, 2014; French & Mareschal, 2017; Mareschal & French, 2017), a recognition-based connectionist recursive autoencoder model of chunking and sequence segmentation that has successfully simulated a significant body of empirical data in the area of syllable- and image-sequence recognition, might also be applied to elementary melody perception. TRACX2 is, indeed, a model of segmentation and chunking, and this article might more appropriately be said to be about "melody *segmentation*", but, in our view, segmentation and chunking are the processes that give rise to perception, hence our title.

The features of early music perception (i.e., in young children) have been the object of study for many years. It is well known that listeners tend to group together similar sounds and, based on regularities perceived in the melodies to which they are exposed, learn to anticipate what will come next. Theoretical frameworks have been proposed to account for these features of the early developmental stages of music perception and various statistical/computational models have been used to simulate them. In the present paper, we will show that a single low-level memory-based segmentation-and-chunking mechanism is able to reproduce some of the basic characteristics of music perception. The work presented here builds on earlier work segmentation-and-chunking in natural language and image processing (e.g., Christiansen, Allen, & Seidenberg, 1998; Cleeremans & McClelland, 1991).

Music perception is more complex than the segmentation and chunking of syllable-streams or image-streams of simple geometric objects, and for this reason, the work in this paper is focused on melody perception, in particular segmentation and chunking as a first, crucial step towards full music perception. The input to TRACX2 consisted of melodies taken from children songs (i.e., a children's songs being coded as melody only, that is, as a sequence of notes), without taking into account the duration of the notes. Timbre, pauses, chords, and emphases, all present in more complex music, were rare in these songs and when they did occur, they were removed. This simplified input comes close to the environment that infants and children actually hear when listening to children's songs (e.g., lullabies, play songs, etc.). TRACX2 is used here to simulate some of the early developmental stages of music learning, in particular melody-related learning. We will show that its internal representations cluster in a human-like manner, that its contour information is also encoded in these representations, and that the ends of motives have a particular importance for the model, as they do for infants. In addition, we briefly compare our model to four other models of sequence segmentation – namely, first-order Markov models, PARSER (Perruchet & Vinter, 1998, 2002), a Recurrent Auto Encoder (RAE, Socher et al., 2011) and a Simple Recurrent Network (SRN, Elman, 1990; Cleeremans & McClelland, 1991).

This article is organized around a series of studies. After a brief summary of what is already known about music perception, we use TRACX2 to simulate four families of studies. We begin by explaining the details of the method used in the simulations, the input data, their internal representations, and the impact of the temporal organization of the tone sets/units (which we refer to as "words", following the tradition of speech segmentation studies, which TRACX2 simulated initially) on the results. We then show that the simple chunking mechanism instantiated by TRACX2 can explain three features of human melody perception – namely:

- melodic motives (defined here as short melodic excerpts of 3 to 4 notes, i.e., 2 to 3 intervals) are identified more rapidly when TRACX2 has already been exposed to other, similar, but not identical, structures. In other words, in TRACX2, as in humans, prior learning improves subsequent recognition of similar items, whether or not they were in the training set.
- TRACX2 is responsive, as are humans, to the melodic contours of the motives it has identified.
- When learning a new melody, TRACX2 recognizes the end of familiar motives better than their beginning, an observation previously reported for humans in statistical learning experiments using melodies/tone sequences.

## 2. Music perception: similarities and differences with syllable-sequence and image-sequence processing

### 2.1. Music perception

At least two different principles have been suggested for how the human auditory system binds discrete sounds together into perceptual units (e.g., Bendixen, Bohm, Szalardy, Mill, Denham & Winkler, 2013): the *feature-similarity principle*, which is based on linking together sounds with similar characteristics over time (temporal proximity, pitch proximity, timbre similarity, etc.) and the *predictability principle*, which is based on linking together sounds that follow each other in a predictable way (e.g., listeners expect upcoming tone-sequences in a melody to be similar to tone-sequences they have already heard either in that particular melody or in general). These principles apply to intervallic differences between notes, to meter, to accents and dynamics, to the consonance of sounds and higher level properties of music linked to tonal structures, such as, the role of the tonic, of other key-defining elements like third and fifth scale degrees, or the equivalence of tones separated by octaves (e.g., Krumhansl, 1983, Schellenberg et al., 2002; Deutsch, 2013). In the simulations presented in this paper, we have simplified the musical material to isochronous melodies and focused on relative pitch, with its intervals and melodic contour.

When tones of different pitch heights are linked together in a sequence, a melody emerges. The differences in pitch height between two adjacent tones (e.g., the tones C and D are separated by two semitones in the upward direction, +2) define intervals, which are the elements of the melodic contour. Contour refers to the pattern of ups and downs of pitch from tone to tone in a melodic sequence. For example, the sequence with the tones C-D-G-E-C-C can be coded in terms of intervals (+2 +5 -3 -4 0), which gives rise to a contour (+ + - - =). Both types of information describe the melody in terms of "relative pitch" information. This means that the melody can be placed at different absolute pitch heights (or be put at different tonal degrees in a given tonal key; Dowling, 1978), while still respecting the same interval pattern and contour (e.g., Dowling & Fujitani,1971). The coding of tone sequences as relative pitch information enables the recognition of a melody regardless of the pitch range of the singer. Even infants can encode tone sequences in terms of relative pitch information by ignoring the change of the pitch range while detecting intervallic changes in the melodic sequence in both short-term and long-term memory tasks (Trehub et al., 1985; Plantinga & Trainor, 2005). Similar patterns have also been observed in adult listeners. For example, in short-term memory recognition tasks in a delayed-matching-to-sample paradigm, performance is better when the "different" item includes a contour change compared to when it preserves the contour (e.g., Dowling, 1978).

Melodic contour has also been shown to play a role in listeners' melodic expectations, allowing them to predict upcoming tone(s) (e.g., Huron, 2006). Narmour (1990) has proposed a theoretical framework for melodies, the implication-realization model, that generates predictions for listeners' expectations. It applies Gestalt principles to the influence of melodic contour (i.e., the patterns of ups and downs) and interval sizes. A just-heard melodic interval

"implies" a certain kind of continuation, and the "realization" of this melodic "implication" allows listeners to integrate the tones into larger melodic patterns. Namour's model contains a set of principles whereby listeners expect future tones to be similar to previous tones, to be proximate, to provide a good continuation, etc. The predictions of Narmour's rather complex model have been tested in a number of experimental contexts (e.g., Carlsen, 1981; Krumhansl, 1995, 1998; Schellenberg, 1996; Unyk & Carlsen, 1987). Results have led to the proposal of reduced versions of the Narmour model that focus on pitch proximity and pitch reversal (Krumhansl, 1995, 1998; Schellenberg, 1996; Schellenberg et al., 2002).

Even though the application of Gestalt principles to music can lead to the hypothesis of an innate, hard-wired basis for music perception, analyses of the statistical distribution of tones also support the hypothesis that listeners can become sensitive to these distributions and features via exposure alone, which then influence melodic expectancy formation (e.g., Huron, 2006). Krumhansl and colleagues applied tone statistics combined with behavioral measurements to the perceptual expectations of listeners for two different musical styles (Finnish spiritual folk hymns, Krumhansl et al., 1999, and North Sami yoiks, Krumhansl et al., 2000). Trained on this data, a Self-Organizing Map (SOM, Kohonen, 1982) suggests that listeners become sensitive to the statistical distributions of tones as well as to higher-order statistics, such as two or three-tone transitions. SOMs are unsupervised connectionist networks that learn regularities in the environment through exposure alone (i.e., without an explicit teacher signal). These networks produce representations of regularities that can be used to simulate listeners' behavior (e.g., in terms of perception, expectations or memory). Krumhansl et al. (1999, 2000) focused on melodic expectations in different styles, while others have used SOMs to simulate tonal knowledge representation with underlying tonal-harmonic relations (Leman, 1995; Griffith, 1995; Tillmann, Bharucha & Bigand, 2000). An SOM, whose connections are shaped by exposure to musical material without a teacher signal, can then be used to simulate empirical data on music perception and memory, as well as tonal expectations (e.g., Tillmann et al., 2000).

Statistical and computational models, as well as various artificial neural networks, have been proposed to describe and simulate human music perception. A significant advantage of connectionist models is their capacity to adapt in such a way that representations, categorizations or associations between events can be learned.

Various other computational approaches have been proposed to simulate musical composition, music performance and improvisation, as well as perception (cf. Cope, 1989; Todd & Loy, 1991; Griffith & Todd, 1999). Music-perception simulations have addressed the perception of timbre, tones, chords and sequences as well as temporal structures. In addition to cognitive simulations, powerful computational models, such as deep neural nets have been used to extract harmonic information from musical audio signals (Korzeniowski, 2018). Other algorithms have been developed in the field of music-information retrieval (MIR) to automatically detect and extract repeated patterns from musical scores (Müller & Clausen, 2007, Nieto & Farbood, 2014) or sound files. However, these latter computational approaches, even though powerful, are unconcerned with the cognitive validity of the procedures and mechanisms used. For cognitive scientists, the simulation of music perception is relevant only insofar as the algorithms used simulate, at least qualitatively, the cognitive processes of the human brain. This includes the generation of errors, confusions, and other problems that arise in real human perception of music, thereby potentially gaining a better understanding of how the human cognitive system processes music.

In the present paper, we adopt this approach and apply a well-known connectionist segmentation-and-chunking architecture (TRACX2) to musical material and, specifically, to melodic processing. This model has previously been successfully applied to the simulation of sequential verbal and visual processing. The extension of the TRACX2 architecture to a new type of material would further strengthen its psychological plausibility as a general segmentation-and-chunking mechanism. That said, it is clear that the model, as well as the simplified, interval information input to it, must be considered merely as a first step in developing statistically driven models (i.e., models that do not include explicit musical rules) of early music perception. One must crawl before one can walk, and it is our hope that this model will provide a jumping off point for future, more sophisticated models based on some of its architectural principles.

## 2.2. Syllable- and image-sequence processing: Similarities and differences to melody processing

TRACX2, and its predecessor, TRACX, have been able to successfully simulate a wide range of experimental data in the area of syllable- and image-sequence data, among them infant data from Saffran et al. (1996a,b), Aslin et al. (1998), Kirkham et al. (2002), Slone & Johnson (2018, two experiments), and French et al. (2011, Equal Transitional-Probability experiment), as well as adult data from Perruchet & Desaulty (2008, two experiments), Giroux & Rey (2009), Frank et al. (2010, two experiments), and Brent & Cartwright (1996). TRACX/TRACX2 have also been shown to be able to generalize to new input and to develop clusters of emergent internal representations that correspond to the clusters of the input data and simulate top-down influences on perception, as observed in the human data sets (French et al., 2011).

### 2.2.1. Similarities

There are a number of obvious similarities between syllable sequences and music interval sequences. A first similarity is linked to the sequentiality of items presented to the system: musical intervals in a melody are processed in a sequential manner. In addition, the sequences, in both visual and auditory modalities, exhibit statistical regularities (non-uniform distribution of the atomic elements, recurring sub-structures, different transitional probabilities from one atomic element to the other, etc.) Furthermore, boundaries exist between chunks of graphical motives or sounds. Atomic elements, and their aggregates have forms that make it possible to define similarities and distances between them, which can be expressed in terms of perceptual distance. Sequence segmentation and chunking require learning. And this learning is particularly sensitive to the closeness of elements, to the adjacency of sounds, syllables, image features. Generalizations to new sequences based on prior learning occur, and prior learning influences new learning.

These similarities suggest that TRACX2 could be an appropriate model for reproducing some of the basic features of melody-sequence processing, thereby hinting at the potential generality of TRACX2's recursive autoassociative chunking mechanism for sequence segmentation and chunking.

### 2.2.2. Differences

There are, however, a number of differences between syllable-sequence, image-sequence and melody-sequence processing. A chunk in a syllable sequence generally corresponds to a "word" in a given language. A chunk in an image sequence generally corresponds to some higher-level image (e.g., a feature or an object). Studies by Saffran et al. (1996a, b) and Aslin et al. (1998) on infant word learning and work by Kirkham et al. (2002), Tummeltshammer et al. (2017) and Slone & Johnson (2018) on image-sequence learning, all start with a pre-defined set of "words" (short syllable sequences or short sequences of geometric images). Long syllable or image sequences are then constructed by concatenating these "words". These sequences of "words" are heard or seen by the infants or adults who are then tested for their capacity to extract these words from the continuous stream. This implementation mirrors processes related to language acquisition, partly based on the segmentation of the speech stream into word units. The same applies for studies on human image-sequence segmentation (Chantelau, 1997). For a given piece of music, however, there is nothing that corresponds to a pre-defined set of sequentially presented tones or sets of tones ("words") out of which the piece of music is built. In a melody, there is generally no such direct correspondence between chunks of notes and clearly recognizable musical "words" (even if in most musical pieces there are highly identifiable motives, like the 4-note opening motif of Beethoven's Fifth Symphony). Nevertheless, "chunks" of frequently-occurring sequences of notes or intervals do fall into certain human-recognizable categories (e.g., a rising interval followed by a descending interval), and listeners are sensitive to this information.

## 3. Computational simulations of melody perception with TRACX2. General methods

### 3.1. General architecture of TRACX2

TRACX2 (French & Cottrell, 2014; French & Mareschal, 2017; Mareschal & French, 2017), and its closely related predecessor, TRACX (French et al., 2011), are recursive connectionist autoencoders (Pollack, 1989, 1990; Blank, Meeden, & Marshall, 1992, Socher et al., 2011) that model sequence segmentation and chunk extraction. The TRACX architecture was originally developed to simulate a pair of classic experiments (Saffran et al., 1996; Aslin et al., 1998) in infant syllable-stream segmentation and chunking. The key features of both the TRACX and TRACX2 architectures (see Figure 1) are as follows:
- it is a three-layer autoencoder (i.e., an autoassociator with a hidden layer) that modifies its weights so that it can reproduce on output what is on its input
- it recognizes "chunks" of sequential items that have been frequently encountered on input;
- it dynamically incorporates the internal representations developed in its hidden layer into new input;
- its internal representations cluster in a manner that is consistent with how the input clusters (i.e., similar chunks have similar internal representations);
- it generalizes well to new input.

The key point about an autoencoder network is that the degree to which its output matches its input is a measure of how often the network has encountered that input before. If it has encountered a particular input often, its output will closely match that input. If, on the other hand, it has not encountered a particular input before, or has encountered it only rarely, there will be a significant error between the input to the network and the output produced by that input.

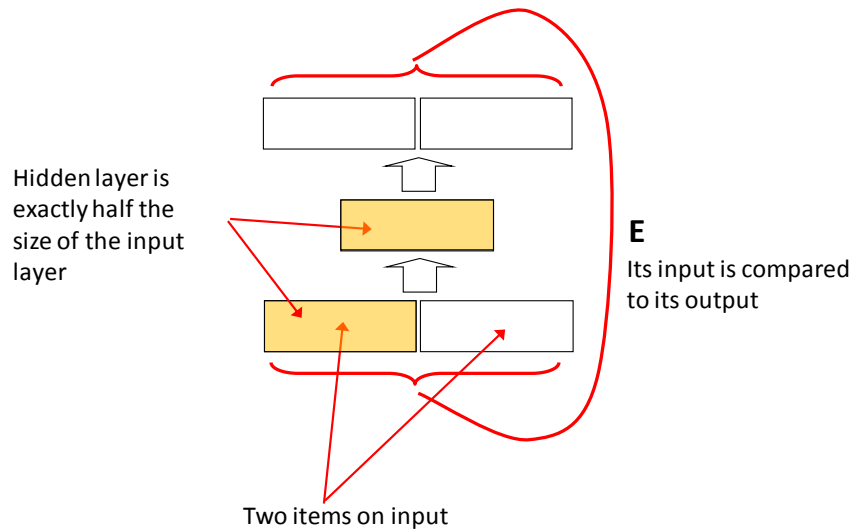

Figure 1. The 3-layer TRACX2 architecture with feed-forward weights between each successive fully connected layer.

The TRACX2 architecture consists of three layers as shown in Figure 1. The input layer is divided into two parts of equal length, the left-hand side (LHS) and the right-hand side (RHS). Crucially, the hidden layer is the same size (i.e., has the same number of nodes) as the LHS and the RHS of the input layer. Bipolar inputs, {-1, 1}, were used. The standard mean squared error function was the objective function used with the backpropagation algorithm.

As with prior simulations using TRACX2, the learning rate of the network was fixed at 0.01 and there was no momentum term. A Fahlman offset (Fahlman, 1988) of 0.1 was used to eliminate flat spots in the learning. The network weights were initialized to random values between -0.5 and 0.5. A bias node was added to the input and hidden layers. A modified ReLU (Rectified Linear Unit) squashing function at the hidden and output layers, which was linear over the interval [-5, 5] and -1 for output less than -5 and 1 for output greater than 5, was used. (A tanh function was used in previous versions of TRACX2. We decided to use a ReLU function because it has become standard practice, especially for deep neural networks, and because it is considerably faster to calculate (Glorot, Bordes, Bengio, 2011) and, finally, it can be adjusted, if need be, more easily than tanh.)

Results for all simulations were averaged over 20 runs of the model with different starting weights of the connection matrices, with the exception of the calculations on the internal representations because combining the network's internal representations over several runs is problematic.

### 3.2. Weight changes

The "teacher" that drives TRACX2's learning is the input itself. In other words, on each weight-change iteration, the network attempts to reproduce on output what was on its input. The difference between the actual output of the network and the input drives the Generalized Delta Rule (Rumelhart & McClelland, 1986), which is used to change the weights of the connection matrices between the layers. A mean distance, defined as the mean of the absolute values of the differences between all of the corresponding values of the input and the output vectors, is used to calculate a dissimilarity measure, denoted by $E$ in Figure 1. To understand the chunking mechanism implemented by TRACX2, we will consider that items: $S_1$, $S_2$, ..., $S_{t-1}$, $S_t$, $S_{t+1}$, ... are sequentially input to the network. At each time step, one new item is put into the RHS of the input. Assume that $S_{t-1}$ and $S_t$ are currently on input. This input, $[S_{t-1}, S_t]$, is fed through the network. This produces a vector, $H_t$, at the hidden layer and a vector on output, [$Out_{t-1}$, $Out_t$], each of whose terms is between -1 and 1. This latter vector is compared to the input vector, [$S_{t-1}$, $S_t$], and a measure of dissimilarity, *E*, between the two is computed. *E* is always between 0 (if the input-output correspondence is perfect) and 2 (if it is as bad as possible). Based on this dissimilarity, *E*, the weights of the connections between the Hidden-to-Output and the Input-to-Hidden layers are changed according to the standard backpropagation algorithm (Rumelhart & McClelland, 1986).

### 3.3. Context-dependent input

What is put on the input of TRACX2 on the next iteration depends on the size of *E*. On the next time step, *t+1*, a weighted combination of $S_t$ and the content of the hidden units $H_t$, is put in the LHS of the input unit : $(1-\Delta)*H_t + \Delta*S_t$ where $\Delta$ is simply *E* squashed by a slightly modified tanh function to be between 0 (when *E* = 0) and 1 (when *E* = 2). A value, which is referred to as Temperature, determines the shape of this modified tanh. The larger the value of Temperature, the steeper this curve. The higher value of this parameter, the steeper the tanh function that determines $\Delta$. In the current implementation we set Temperature = 5. The system also puts the next item, $S_{t+1}$, in the sequence into the RHS of the input. This means that if $\Delta$ is close to 1 (as is the case at the beginning of learning, when the difference between the network's input and what it produces on output is high), the network essentially slides item $S_t$ from the RHS of the input to the LHS, (and puts $S_{t+1}$ into the RHS of the input). If, on the other hand, $\Delta$ is very small, the network "assumes" that it has seen the input pair [$S_{t-1}$, $S_t$] many times before (which is the only way $\Delta$ could be very small). Any pair of inputs that occur together many times is considered by the network to constitute a "chunk", which is encoded by the hidden units, $H_t$. Thus, on the next iteration (i.e., at time *t+1*) the network puts, not $S_t$, but essentially $H_t$, TRACX's hidden-unit representation of the chunk [$S_{t-1}$, $S_t$] into the LHS of the input, and then, as before, puts $S_{t+1}$ (the next incoming item) into the RHS of the input. When $\Delta$ is neither large nor very small, the content at t+1 of the LHS of the input is a mixture of the internal representation, $H_t$, and of the preceding RHS, $S_t$ (Figure 2).

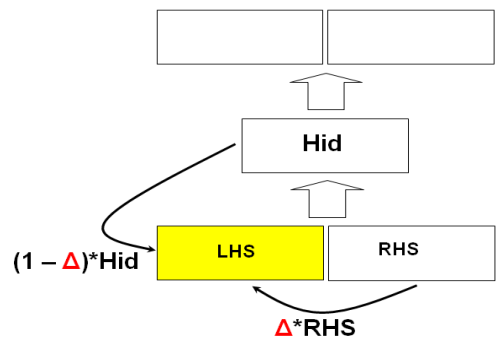

Figure 2. The architecture of TRACX2. (Hid refers to Hidden units. LHS/RHS to the Left-hand side/Right-hand side of the input layer.)

In this way, the network chunks frequently-seen pairs of input and re-uses those chunks dynamically to potentially create increasingly larger chunks from the input. Assume, for example, that the item sub-sequence, *abc*, is a frequently repeated subsequence in the item sequence. At some point, the pair, *a-b*, on input (*a* in the LHS and *b* in the RHS of the input) would be recognized as having been seen together often and *E* would become small. Therefore *a-b* will be considered to be a chunk by the network. TRACX2's internal representation (i.e., hidden-layer representation) of *a-b* would be *H(ab)*. So, on the next time step, essentially *H(ab)* plus a very small contribution from *b*, rather than only *b*, would be put into the LHS of the input and, as always, the next item in the sequence, in this case, *c*, would be taken from the item sequence and put into the RHS of the input. Once the input pair [*H(ab), c*] produced output that closely resembled the input, [*H(ab), c*] would be chunked as *H(abc)*. In this way, larger and larger chunks of items, if they occur together frequently in the item stream, will be chunked by TRACX2.

It is important to note at this stage that if Δ is always given a value of 0, the network will function as a Recurrent Auto Encoder (RAE). In the present paper, we have extensively compared the behavior of TRACX2 to both the RAE and an SRN.

### 3.4. Input data

We trained TRACX2 on a series of well-known French children's songs in which only pitches are considered (all with equal duration). These songs (Set 1) were: *Ah les crocodiles; Bateau sur l'eau; Fais dodo, Colas mon p'tit frère; Au clair de la lune ; Ainsi font; Une souris verte; Ah vous dirai-je maman; Pomme de reinette; Sur le pont d'Avignon; Frappe, frappe, petite main*. To ensure that our results were not dependent on our choice of children songs, we also trained TRACX2 on a second set of similar children's songs (Set 2): *Alouette, gentille alouette; Biquette ne veut pas sortir du chou; Dans la forêt lointaine; Je te tiens, tu me tiens; Le bon roi Dagobert; Il était une bergère; J'ai du bon tabac; J'ai perdu le do; Frère Jacques; Il court le furet*. Features, such as rhythm, meter, tempo, harmony, and texture were not taken into account. Based on the importance of relative pitch, intervals and melodic contour in music perception, for all of the simulations reported in this paper we encoded, not notes, but rather the intervals between notes. So, just as the "primitives" in Saffran et al. (1996a, b) and Aslin et al. (1998) were individual syllables, the primitives in Slone & Johnson (2018) were a small number of the geometrical shapes (e.g., crosses, triangles, circles), and the primitives in Saffran et al. (1999) were musical notes, the primitives of our simulations were the intervals between successive notes. The difference with respect to the above studies, of course, is that we did not construct the melodies used from our set of primitives.

In order to test a possible prior-learning effect of the network's exposure to these children's songs, we used the first 42 measures of the Allegro Assai of the Sonata for Violin Solo in C Major BWV 1005 by J. S. Bach.

The children's songs and the part of the Bach sonata BWV 1005 that we used in the prior-learning study required a total of 39 intervals (Figure 5b). (The children's songs contained only 25 of these intervals.) Figure 3 shows a short melody with labels of pitch and the intervals between pairs of tones. From the note *A* to the note *E*, there is a decrease in pitch by 5 semitones, hence an encoding of -5. Between *E* and *B*, on the other hand, there is a rise of 7 semitones, thus +7. (Figure 3).

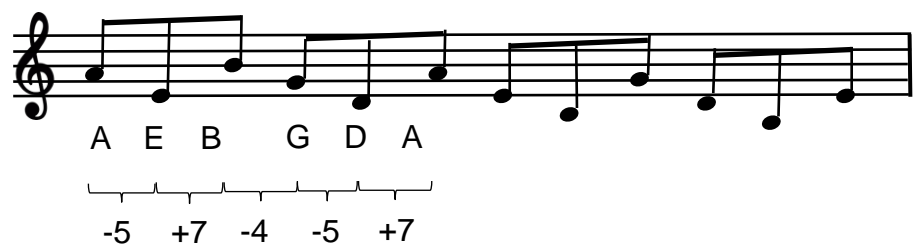

Figure 3. Encoding of intervals between notes. The number indicates the number of semitone steps between the notes and the +/- sign the direction (+ up or - down).

Figure 4 shows how these intervals were labeled for the purpose of the present simulations. For convenience and for accessibility for non-musician readers, we labeled each of the intervals from -19 to 19 with capital letters included or from *a* to *y,* instead of using the music-theory terms, with + or -  indicating the direction (rising or falling) of the interval. Two-note intervals are designated in our paper here by capital letters or lowercase letters: *A,*

*B, ... a, b, c, ..., x, y, ... Y, Z*. There were only 25 intervals (from -12 to +12) in the children's songs, and these were labeled with lowercase letters from *a* to *y*. (See Figure 4)

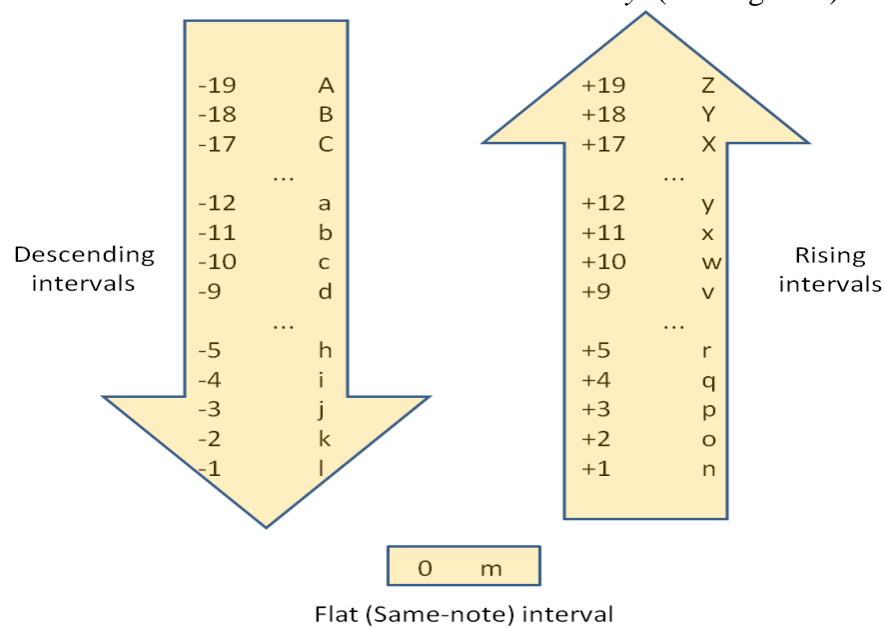

Figure 4. The labeling of the 39 different intervals found in the children's songs and in the Bach sonata.

Two types of encoding were tested. We initially used a one-hot encoding scheme, where each interval was represented by a single unit set to 1 with all others set to -1. This type of encoding was used by TRACX and TRACX2 when simulating segmentation and chunking of syllable- and image-sequences (French, 2011; Mareschal & French, 2017). But we rapidly realized the limitations of that scheme for encoding musical sequences. Unlike for syllables and geometrical images, there was a clear need to impose a distance metric on the input coding of intervals. In a musical piece, the passage from the tone C to the tone D is perceived as being very different than going from C to A, something one-hot encoding cannot capture. The first pair describes an upward movement with a distance of two semitones (+2), whereas the second pair describes a downward movement with a distance of three semitones (-3). Depending on the pitch-height difference of the two tones, the pairs/contours described in this manner have greater or lesser perceptual similarity. Our study of TRACX2's internal representations after learning, for example, clearly showed the necessity of maintaining the proximity information of the intervals input to the network. We, therefore, replaced the traditional one-hot encoding by what we called an "ordinal" encoding of the input (also called "thermometer encoding" in the machine-learning literature). In addition, the error measure, $E$, used to drive TRACX2's backpropagation learning had to be adapted to this type of input encoding: the original Chebyshev distance (a maximum distance) used in prior versions of TRACX and TRACX2 had to be replaced by the mean absolute difference between the input and output vectors. Finally, we made the assumption that listeners are perfectly able to discriminate the here used tone differences. This is a reasonable assumption as the required minimal discrimination between two tones here was a semitone apart (i.e., +1 or -1), and pitch discrimination thresholds for non-musicians have been reported to be inferior to a semitone (e.g., an average of 0.22 semitones, Pralus et al., 2019).

The ordinal encoding of the musical intervals encountered in the set of children's songs was done as follows:
    A  1, -1, -1, -1, -1, -1, ..., -1
    B  1, 1, -1, -1, -1, -1, ..., -1
    C  1, 1, 1, -1, -1, -1, ..., -1

```
...
X  -1, ..., -1, -1, -1,  1, 1, 1
Y  -1, ..., -1, -1, -1, -1, 1, 1
Z  -1, ..., -1, -1, -1, -1, -1, 1
```

Ordinal encoding reflects both the size and direction of the intervals. So, for example, *m* is the interval corresponding to the repetition of a note and, therefore, has a value of 0, *o* is the interval corresponding to a rise in pitch of 2 semitones, and *t* corresponds to a rise of 7 or a perfect fifth. Ordinal encodings of *m* and *o* differ by two bits, whereas *m* and *t* differ by seven bits. The use of ordinal encoding corresponds, or at least approximates, what a human would perceive in listening to *m* and *o* versus *m* and *t*.

### 3.5. Procedures used for training and testing

The entire training corpus of children's songs was presented to the TRACX2 network for 30 epochs. We chose this small number of epochs compared to the many thousands of epochs generally used in connectionist networks, in an attempt to simulate, in a very approximate and conservative manner, the number of times a young child might be exposed to these songs. On each training epoch the order of the songs presented to the network was randomized. It is clear that many children listen to these songs considerably more than 30 times, but our aim was to avoid typical connectionist training regimes of many thousands of epochs, since it is not clear what these enormous numbers of training cycles actually correspond to empirically. We, therefore, chose a small number of epochs to model the data, even though this might seem unusual in comparison to standard connectionist simulations.

### 3.6. Description of the training data

The distribution of all the intervals contained in the two sets of children songs is shown in Figure 5a. By far the most frequently encountered interval was the one in which the two successive notes are identical. This contrasts with an excerpt of a Bach sonata that constituted one of our test pieces, in which there were no such intervals (Figure 5b). In addition, in the children's song corpus less consonant intervals, such as tritones (e.g., *s* = +6) and minor sixths (*u*, +8), were completely absent.

The first set of 10 songs used to train the network contained a total of 437 intervals. During training there was no intervallic connection between the last interval of one song and the first interval of the next song. The average size (measured in semitones) of the 437 intervals was -0.0092. In other words, ascending (+) intervals nearly exactly balanced out descending (-) intervals. Their standard deviation (measured in semitones) was 3.45.

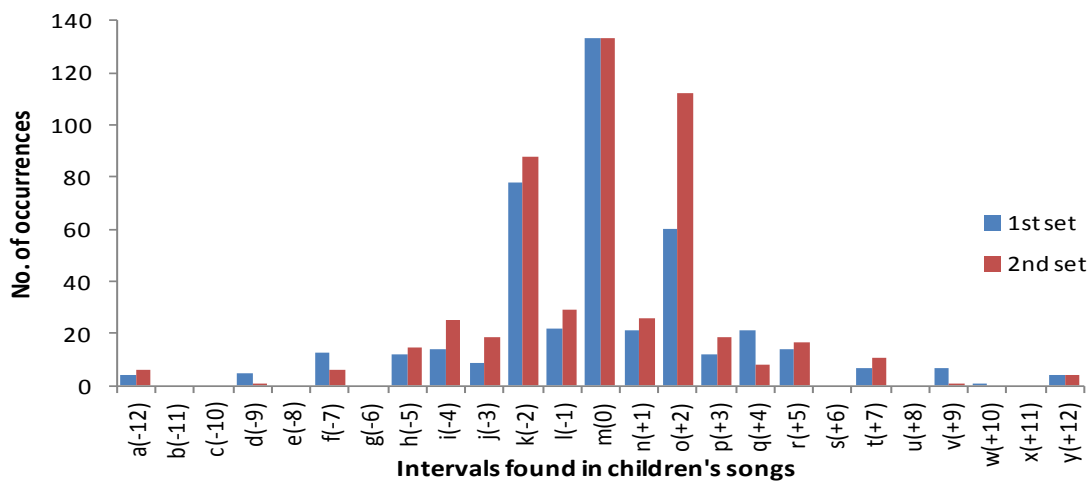

Figure 5a. The distributions of all intervals encountered in the two training corpora of children's songs (Set 1, Set 2).

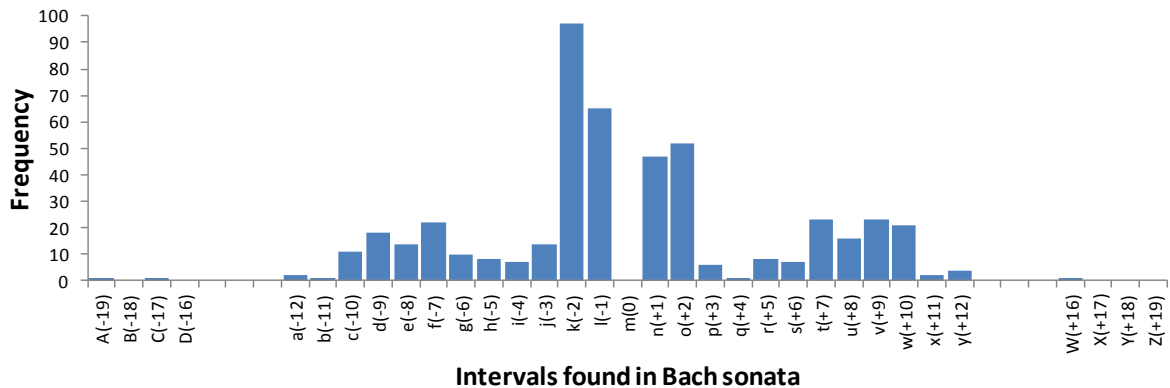

Figure 5b. Distribution of intervals in the first 42 measures of Bach's sonata for violin BWV 1005. The size of the intervals is indicated on the x-axis. Note the complete absence of the "flat" interval, *m,* of size = 0.

We also analyzed the distribution of all 2- and 3-interval "words" in the training corpus. In keeping with the literature on sequence segmentation (e.g., Saffran et al., 1996a, b; Aslin et al., 1998; Slone & Johnson, 2018), we have called short sub-sequences of intervals "words" instead of using terminology like bigrams, trigrams or triplets. Figure 6 shows this distribution for 2-interval words in Set 1 of the children's songs

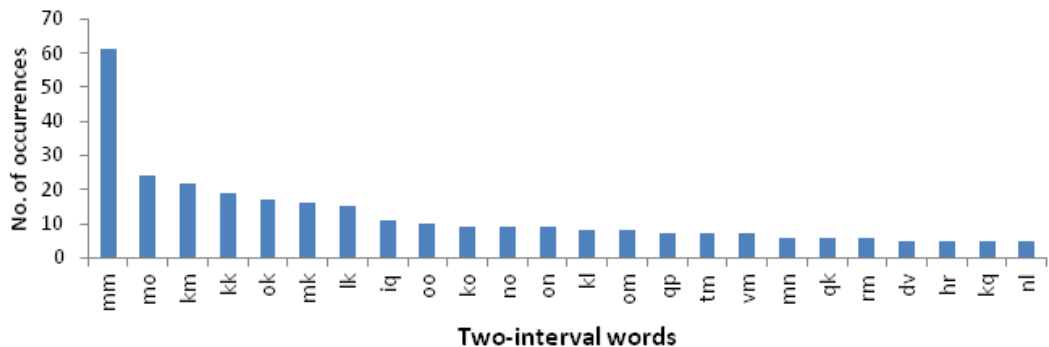

Figure 6. Raw frequencies of 2-interval words appearing in the first training corpus at least 5 times.

*mm* words (i.e, words consisting of two occurrences of the "flat" interval, *m*, in which both notes are identical) are, by far, the most frequently encountered 2-interval words in the training corpus. For 3-interval words, *mmm* was the most common. (These distributions are essentially the same when the program is run on the second set of children's songs.)

### 3.7. Word error calculation

The degree to which TRACX2 recognizes words is based on the input-output error produced when a word is presented to its input. For 2-interval words, the word is encoded and input to the network. Activation then spreads via the hidden layer to the output and the error value, *E*, is calculated, as indicated earlier, as the mean absolute difference between the input and output vectors. A small error means that the word is well recognized by (i.e., is "familiar" to) the system, whereas a large error means the word is not well recognized by the network, because it is new or has been seen only infrequently by the network.

For 3-interval words, the error-calculation is somewhat more complex and will be explained by means of a concrete example. Consider the 3-interval word, *kmm*. At time *t*, the interval *k* is put on the LHS of the input, and *m* on the RHS. *km* is then fed through the network and the output error, $E_1$, is calculated. $E_1$ is then converted by a modified tanh into $\Delta$ (see Section 3.3 *Context-dependent input*), which then determines how much of the hidden-unit activations and the RHS activations at time *t* are to be included in the LHS of input at time *t+1* (see Figure 2). In other words, as was done during the original learning of the first two intervals, the LHS is filled with a combination of the hidden-unit vector ($H_t$) plus the RHS input vector -- specifically, $(1-\Delta)*H_t + \Delta*RHS$. The encoding of the second *m* is then put into the RHS of the input vector. This full input is then fed to the output nodes of the network, and the mean absolute error between input and output ($E_2$) is calculated. The average of $E_1$ and $E_2$ is used as the error-measure for *kmm*.

## 4. Study 1: TRACX2's internal representations

In this section, we examine TRACX2's internal representations. We address the following question : What kind of information is encoded in TRACX2's internal representations and how is that information organized? Three different studies (St1.1, St1.2 and St1.3) will be considered.

In the original TRACX paper, French et al. (2011) showed that the internal representations of TRACX clustered in a way that tracked the grammatical structure of the syllable sequences that were input to it. Do we get similar results and do TRACX2's internal representations create clusters of similar musical 2-interval words? St1.1, therefore, looked at the "topological organization" of TRACX2's internal representations.

We then decided to examine the internal representation of longer words. St1.2 studied whether the network keeps a trace in its internal representations of the values of the intervals that define these longer words.

Finally, we studied (St1.3) the relationship between the errors of words and their temporal location in the training set. In particular, we investigated whether there are primacy or recency effects.

### 4.1 General method
In the three studies, we considered the internal representations and the errors that TRACX2 generated after training on the children's songs. The simulations were done on both the primary set (Set 1) and the verification set (Set 2) of children's songs, and the results were essentially identical. We present the results of a number of simulations carried out by TRACX2 (Figure 8) and compare the performance of the model with other systems -- namely, first-order Markov chains (i.e., transitional probabilities only), PARSER, an RAE and an SRN.

### 4.2. PCA grouping of 2-interval word contours (St1.1)
4.2.1 Method

We trained TRACX2 on one set of children's songs. We then performed a principal components analysis of the first two principal components of the 39-element hidden-unit representations of all of the 84 2-interval words in the training corpus. The various types of contours of 2-interval words can be defined depending on whether their component intervals were rising (R), falling (F), or flat (=). In all, nine clusters of two-interval contours will thus be considered: rising-rising (RR), flat-rising (=R), falling-rising (FR), falling-flat (F=), falling-falling (FF), flat-falling (=F), rising-falling (RF), rising-flat (R=), and, finally, flat-flat (==).

### 4.2.2 Results

Unsurprisingly, no reasonable clustering of the hidden-unit representations was obtained when we used one-hot coding for the intervals input to the network. The points projected onto the plane of the first and second principal components did not cluster according to their contour. However, when ordinal coding was used, we discovered that the internal representations of TRACX2 cluster in a very meaningful way (Figure 7).

Figure 7 ("TRACX2 contours") shows the space defined by the first two principal components of the hidden-unit representations of the 84 2-interval words found in the first set of children's songs. This figure clearly shows that 2-interval words with similar contours tend to group together. It is interesting to note that the clusters containing a flat interval are exactly where they should be with respect to the larger clusters on either side of them. For example, consider RR (R means "rising"), the cluster of hidden-unit representations of 2-interval words where both intervals are rising, and RF (F means "falling"), the cluster where the first interval rises and the second falls. R= (= means "flat") is the cluster representations of words whose first interval rises (like RR and RF) and whose second interval is flat (i.e., "between" rising and falling). In other words, the R= cluster should reasonably fall between the RR and RF clusters, which, in fact, it does. The same is true for all of the other clusters containing a flat interval.

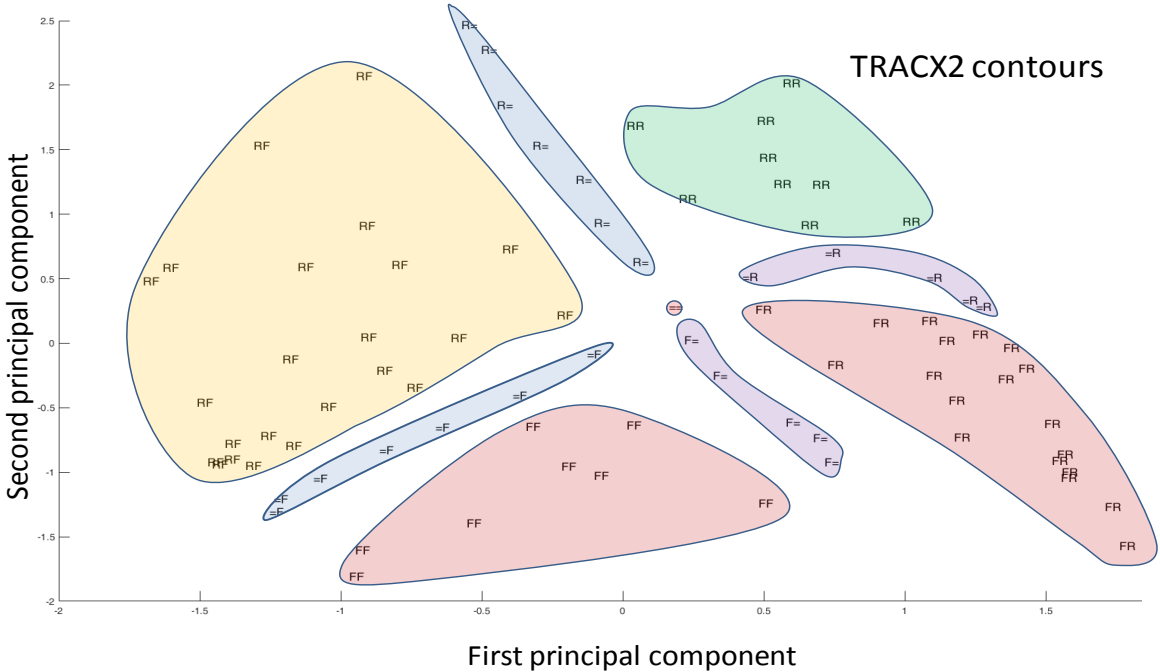

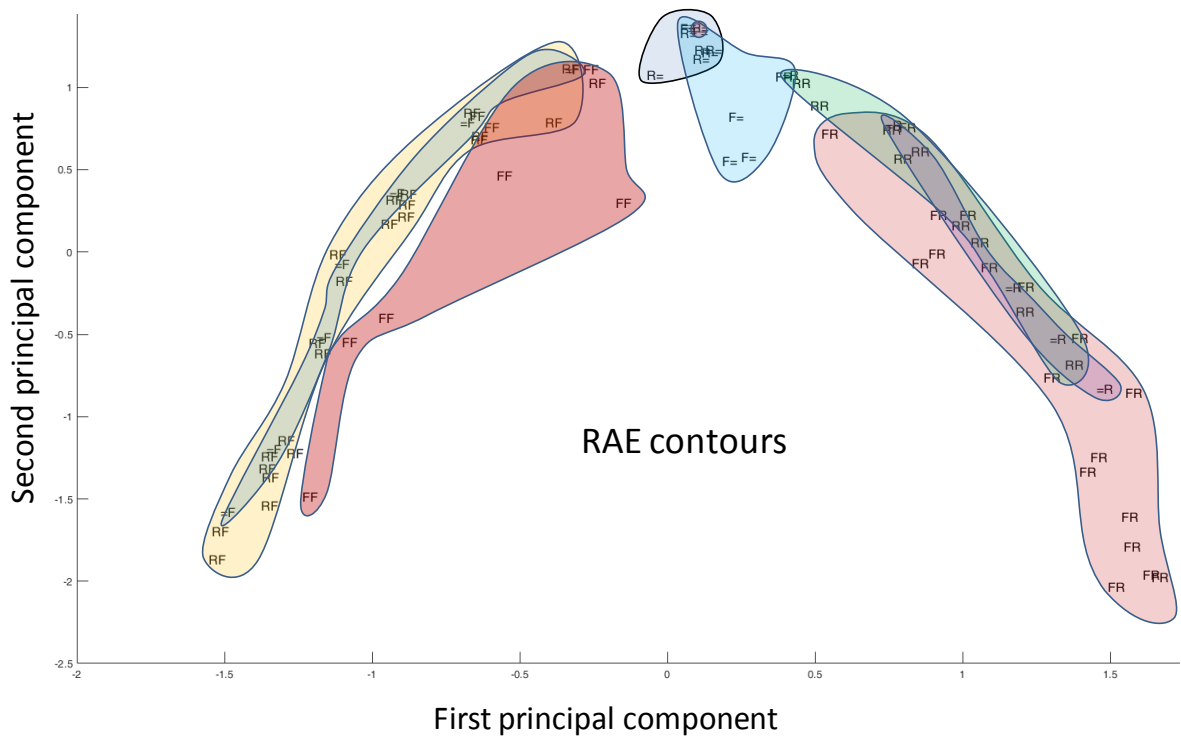

| | |
|---|---|
| RR: Both intervals are rising | FF: Both intervals are falling |
| RF: 1st interval is rising, 2nd is falling | FR: 1st interval is falling, 2nd is falling |
| R=: 1st interval is rising, 2nd is flat | =R: 1st interval is flat, 2nd is rising |
| F=: 1st interval is falling, 2nd is flat | =F: 1st interval is flat, 2nd is falling |
| = =: Both intervals are flat | |

Figure 7. Comparison of TRACX2's and RAE's clusters of internal-representation of 2-interval-word contours after 30 epochs of learning of the children's songs.

In addition, for TRACX2, within each class of intervals containing the flat interval, *m*, and a rising or falling interval (i.e., R=, =R, F=, and =F), the distance of each word in the class from *mm* (the word with two flat intervals, ==) depends on the size of the rising or falling interval making up the word. To see this, consider the clusters F=, made up of the words: {*fm, hm, im, km, lm*} and R=, made up of the words: {*nm, om, pm, qm, rm, tm, vm*}. The sizes of the two intervals making up each word are shown in square brackets. Starting at *mm* (i.e., "==") and moving downward through the class, F= consists of, in order: {*lm* = [-1,0], *km* = [-2,0], *im* = [-4,0], *hm* = [-5,0], and *fm* = [-7,0]}. Starting at *mm* (i.e., "==") and moving upward through the class, R=, consists of, in order: {*nm* = [1,0], *om* = [2,0], *pm* = [3,0], *qm* = [4,0], *rm* = [5,0], *tm* = [7,0], *vm* = [9,0]}

Thus, it can be seen that distances and directions from "==" (the "flat word") correspond precisely to the size and +/- direction of the non-flat interval in each of the words in these two classes. The same is true for the classes **=R** and **=F**.

### 4.3. Does the memory trace of longer words contain traces of its components? (St1.2)
4.3.1. Method

We carried out analyses on longer words to see whether a trace is kept in the internal representations of the values of the successive intervals that made up the words in the children's songs. An example illustrates the method we used. Consider the 4-interval word *mnoh*. It can be characterized in two different ways:
  i) from the values of its 4 intervals, namely  0 (*m*), +1 (*n*), +2 (o) and -5 (*h*). We will denote these 4 values by $I_1(mnoh)$, $I_2(mnoh)$, $I_3(mnoh)$, $I_4(mnoh)$, respectively;
  ii) from its internal representation, denoted by R(*mnoh*), a vector of 39 real numbers.

Consider $I_1$. It is the function that associates the 4-interval word *mnoh* with the value of its first interval, i.e., $I_1(mnoh) = 0$. As training and chunking progress, *m* is first chunked with *n*, then *mn* is chunked with *o* and, finally, *mno* is chunked with *h*. This means that the interval *m*, as such, has progressively disappeared as a distinct input to the network. But is its value retained in one way or another in the internal representation, R(*mnoh*)?  In other words, can we reconstruct $I_1$ from R? And are $I_2$, $I_3$ and $I_4$, also "hidden" in R(*mnoh*).

A simple way to determine the extent to which $I_1$, $I_2$, $I_3$ and $I_4$ are "present" in R(*mnoh*) is to calculate the multiple correlation between $I_1$ and R (as well as $I_2$, $I_3$, and $I_4$, respectively, with R). If this correlation is high, then the value of $I_1$ can be derived as a linear combination of the components of R, which means that it can be reconstructed from R.

4.3.2. Results

The analysis carried out on longer words showed that, with ordinal encoding, a trace was kept in the internal representations of the values of the successive intervals that made up the words in the children's songs. With one-hot coding this trace was much poorer (see Table 1), which is one of the main reasons that we rejected one-hot coding for modeling early melody perception.

The table below gives the values of the multiple correlations for both one-hot and ordinal encoding for words of length 3 and 4. We show in parentheses the values obtained on the second set of children songs.

3-interval words

| Coding | Multiple square correlation with $I_1$ | Multiple square correlation with $I_2$ | Multiple square correlation with $I_3$ |
|---|---|---|---|
| One-hot | 0.44 (0.38) | 0.63 (0.45) | 0.93 (0.90) |
| Ordinal | 0.97 (0.98) | 0.98 (0.96) | 1 (1) |

4-interval words

| Coding | Multiple square correlation with $I_1$ | Multiple square correlation with $I_2$ | Multiple square correlation with $I_3$ | Multiple square correlation with $I_4$ |
|---|---|---|---|---|
| One-hot | 0.23 (0.26) | 0.36 (0.32) | 0.57 (0.41) | 0.84 (0.80) |
| Ordinal | 0.84 (0.83) | 0.85 (0.91) | 0.97 (0.97) | 1 (1) |

Table 1. Multiple square correlations of subwords making up longer words (3- or 4-interval words)

Clearly, ordinal encoding enables the system to keep a trace of the components making up its internal representations of the whole structure of the words. As expected, the trace decreases with the length of the word. The final interval of a word is better memorized than the first one.

These results go some way in demonstrating a "chunking" effect at the level of non-adjacent dependencies. The fact that both I1 and I3, for 3-interval words, and I1 and I4 for 4-interval words, have a high multiple correlation with the internal representation means that the system establishes through its internal representations a link between non-adjacent intervals. However, to show that TRACX2 is explicitly sensitive to non-adjacent dependencies would require additional analyses as chunks are progressively built from co-occurrences of adjacent elements. This is clearly an issue that should be explored in future work.

**4.4. Word errors and their relation to frequency and order of appearance in the training set (St1.3)**

4.4.1. Method

We examined the errors associated with the 2-interval words and their relation to their frequency and order of appearance in the training sets. To investigate the possibility of a primacy effect, we take the musical sequence obtained by chaining the 10 different songs (no break between the songs) to get a new input of length 437. After training the network on that sequence, we obtain new errors. They are compared with the previous ones (generated with the 10 songs in the way described in section 3). We also modified the sequence in different ways by moving occurrences of some words to the beginning of the sequence. This was done to test the possible effect of the order of appearance of those words on their associated errors.

4.4.2. Results

Errors associated with 2-interval words are negatively correlated with their frequencies ($r = -0.35$). A word that has been seen by the system frequently generally will have a smaller error on output than infrequently encountered words. However, for certain words in the training corpus, this is not the case. For example, the 2-interval word, *mi*, has a low error on output (0.16), even though it occurs relatively infrequently in the training corpus (only 4 times). On the other hand, the more frequent 2-interval word, *ok*, has a high frequency of occurrence of 17 but, nonetheless, has a higher output error (0.19) than *mi*.

This apparent discrepancy is due to the impact of the temporal organization of words. A close look at the songs in the training set shows that *mi* occurs *at the beginning* of one of the 10 children's songs, thereby potentially producing a primacy effect. After training the network on a new sequence obtained by concatenating the 10 different songs, the error associated with the high-frequency word *mo*, which appears 24 times in the sequence, was 0.25. However, by moving all 24 occurrences of *mo* to the beginning of the 437-word sequence, the error associated with *mo* dropped to 0.15. These results are in line with the well-known primacy effects in memory tasks reported by Ebbinghaus (1913). They are also similar to those reported in a study by Deliège (2001) that demonstrated improved memory performance for first-heard cues in music-recognition tasks.

**4.5. Comparisons with other other models**

4.5.1. Comparison with First-order Markov Chain (TP) calculations

Other statistical learning mechanisms have been shown to be able to extract words from sequences of syllables, and these mechanisms also apply to other domains (e.g. Christiansen et al., 1998; Cleeremans & McClelland, 1991). In most existing models based on statistical regularities, chunks and boundaries between chunks are detected by the variation of transitional probabilities. For example, word boundaries fall where inter-syllable transitional probabilities (TPs) are significantly lower than the preceding and following TPs.

For all 2-interval words in the primary set of children's songs (Set 1), we computed the Transitional Probabilities (TPs) from the first interval to the second. If TRACX2's errors for these words were closely correlated with these TPs, it would be reasonable to claim that the mechanisms instantiated in TRACX2 could have been achieved by simple statistical first-order Markov chain estimations.

To investigate that assertion, we computed the Pearson correlation coefficient, $r$, between the TPs and the errors obtained with TRACX2 on the 84 2-interval words making up the primary set of children's songs (Set 1).. Large errors (i.e., poor chunks) should correspond to low TPs. However, this is not the case. The value of $r$ was 0.13 (i.e., positive and close to 0). In short, errors calculated by TRACX2 for these words did not depend linearly on the associated TPs.

These results have also been confirmed with analyses on 3-interval words, which correspond to words comprising 4 tones. For these words, we replaced simple TPs by average transitional probabilities. The Pearson correlation of the average TPs with the errors made by TRACX2 (see section 3.7. to see how errors on 3-interval words are calculated) on the 161 3-interval words in the primary set of children's songs was found to be 0.32.

In short, TRACX2's errors seem to be capturing not only TPs, but also other types of statistical regularities in the songs.

### 4.5.2. Comparison with PARSER

PARSER (Perruchet & Vinter, 1998, 2002) is a largely, if not completely, symbolic model of syllable sequence parsing and chunking. A particularly clear description of this model can be found in Perruchet & Vinter (1998). It does not maintain anything that is equivalent to the internal representations of TRACX2, aside from what is stored explicitly in its Working Store. For instance, it has no way of knowing that the two-interval FR contour *ay* (12-note fall, followed by a 12-note rise) should cluster with the much less "severe" FR contour *ko* (2-note fall, followed by a 2-note rise), rather than with *ya* or *ok*, both RF sequences. In other words, PARSER was not designed to cluster representations of its data, and hence there is no clustering of musically similar 2-interval pairs.

### 4.5.3. Comparison with RAE

As mentioned above, an RAE is a special case of the more general TRACX2 architecture. Given the simplicity of the RAE, it is interesting to contrast its behavior with the results obtained with TRACX2, parameterized as described in this paper. With an RAE, the projection of the points representing the 2-interval words on the first principal plane after 30 learning epochs is very different from the one obtained with TRACX2 as shown in Figure 7.

We then looked at the correlation between the mean errors over all 84 2-interval words for TRACX2 and RAE. This was 0.79. However, the means (of these mean errors) were for TRACX2 and RAE 0.17 and 0.50, respectively. In other words, the overall errors-on-output (i.e., the fit-to-data) produced by TRACX2 were three times better than those for RAE. (And this difference held up for 3 and 4-interval words.)

We also calculated the correlation between the errors made by TRACX2 and the difference "error RAE – error TRACX". The value is -0.41 for 2-interval words and goes to -0.85 for 3-interval words and to -0.92 for 4-interval words. This means that when TRACX2's error is small (i.e., for familiar words), the difference with RAE is big, RAE errors being larger than TRACX2 errors.

Chunking in TRACX2 and RAE works more or less in the same way. Words producing large errors (non-familiar words) are basically the same for the 2 systems. This is also the case for words with small errors. However, the *differences* between the errors made by TRACX2 and those made by RAE increase as the words become less familiar to the two systems. This could be explained as follows. When words are familiar, TRACX2 and RAE work in a comparable manner. For familiar words on input, the left part of TRACX2's input is mainly the internal representation of the first part of the word. But for RAE, regardless of whether the words on input are familiar or not, it *always* puts their internal representation on input on the next time step. For this reason, for non-familiar words the principles of functioning of the two systems are different. Consider an unfamiliar 3-interval word. RAE takes as input the internal representation of the first 2 intervals, *even if they do not constitute a chunk*. This will then produce a larger error on output than for TRACX2 because, in this case, TRACX2 does not use the internal representation of a non-existing chunk.

### 4.5.4. Comparison with SRN

Given its importance in similar studies (see French et al., 2011, for details), we ran a vanilla SRN (Elman, 1990) on the primary set of children's songs with 30 learning epochs and compared the errors[1] for each of the 2-interval pairs of this set with those produced by TRACX2. Insofar as possible, we set the parameters of the SRN, such as its learning rate, momentum, number of hidden units, Fahlman offset (Fahlman, 1988), number of learning epochs, and its mean absolute error measure to be the same as those used by TRACX2. In spite of these similarities, it is worth mentioning that the tasks of the SRN and of TRACX2 are fundamentally different -- namely, the SRN tries to predict the upcoming interval and TRACX2 tries to reproduce the input.

As for TRACX2 and RAE, we looked at the first two components of a principal-components analysis (PCA) of the internal representations of the SRN for the 84 2-interval words in Set 1 of the children's songs. The clusters of the contours of these words closely resembled those produced by RAE, in particular, with a great deal of overlap. This is not particularly surprising, given that the "context units" at time *t* of an SRN are a copy of the hidden-unit activations of the network at time *t-1*, which is the same mechanism used on input by the RAE.

Finally, we found a correlation of 0.31 between the errors generated by TRACX2 and those produced by the SRN. The reason this correlation is not higher is because of the way in which the 2-interval words are learned. This is illustrated by two relatively infrequent 2-interval words, *ay* (4 occurrences) and *dv* (5 occurrences), compared to high-frequency words, such as *mm* (61 occurrences), *km* (24 occurrences) or *ok* (17 occurrences). These low-frequency pairs were close together in the training set (thus, rapid reinforcement during learning) and had transitional probabilities of 1. This meant that for SRN *ay* and *dv* were among the best learned words, whereas TRACX2, which relies on their frequency of occurrence rather than their transitional probabilities, they were among the most poorly learned words**.**

## 5. Study 2: The effect of prior learning on recognition performance of previously unseen words

---

[1]To calculate the error produced by an SRN for a particular word means setting the context units to 0 and sequentially inputting the items making up the word to the SRN. Setting the context units to 0 is justified because of the distribution of intervals in the children's songs. Because the ascending (+) intervals almost exactly balanced out the descending (-) intervals in the training corpus (Figure 5a), it is reasonable to start with an activation in the context units of 0. The output error is then the average of the prediction errors associated with each of the items making up the word.

Can TRACX2 generalize its learned representations of musical chunks to new, unobserved interval sequences? We will present the results of a number of simulations carried out by TRACX2 (Figure 8) and compare the performance of the model with other systems -- namely, first-order Markov chains (i.e., transitional probabilities only), PARSER, RAE and an SRN. The study is composed of two parts. First, we examined the effect of modifying the familiarization corpus and in a second set of simulations, we examined the effect of prior learning on three different kinds of test items.

**5.1 The effect of modifying the familiarization corpus**
5.1.1. Method

We trained TRACX2, RAE and an SRN on four different, but related training sets. These were the primary set of children's songs and three other sets in which the intervals of these children's songs were scrambled in different ways. For each network, we also included a fifth simulation where there was no prior learning. After training the networks on these different versions of the primary data set (and without training), we selected a set of 3-interval words that did not occur in any of the training corpora, but that were found in the Bach sonata. We called this set the "Bach test words". Each of the following training/test procedures was run 20 times, each time reinitializing each network's weights.

All networks were trained for 30 epochs (with the standard values of learning rate, momentum, etc., See Section 3.1) on the primary corpus of children's songs ("songs" in Figure 8). We then fixed the weights of the networks and presented the Bach test words to each network and recorded the errors obtained.

To see the role played by the intervals themselves, independently of their order, we then randomly permuted the intervals in each of the children's songs ("within-song permute" in Figure 8), and, starting with newly initialized, random weights, trained the networks for 30 epochs on these scrambled children's songs. We fixed the networks' weights and tested their recognition performance, as measured by errors on output, on the Bach test words.

We also created a third training corpus by randomly distributing all of the intervals across all ten of the children's songs of the primary set ("global permute" in Figure 8). This was intended to test a possible, more general familiarity effect with intervals frequently encountered in the children songs. After randomly re-initializing each network's weights, we trained them for 30 epochs on this corpus, fixed their weights and tested each network's recognition performance on the Bach test words.

We then randomly chose intervals from the full set of 39 intervals and distributed these intervals across all ten of the children's songs ("full random permute" in Figure 8). This last simulation was designed to test a possible learning effect on musical intervals, a kind of by-product of the general learning mechanism used in neural networks. As before, we reinitialized all of the networks' weights, trained them on this set, fixed their weights and tested their recognition performances on the Bach test words.

Finally, after once again re-initializing the networks' weights, we tested each network on the Bach test words with no prior training.

In each case, the length of each song (i.e., the number of intervals) was left unchanged.

We also ran these tests for the RAE and the vanilla SRN, as described above (Figure 8).

We averaged our results over the 20 runs of the program for each of these training/test scenarios. In all cases, we used the standard set of learning parameters for TRACX2, the RAE and the SRN.

5.1.2. Results

The results of the simulations are shown in Figure 8.

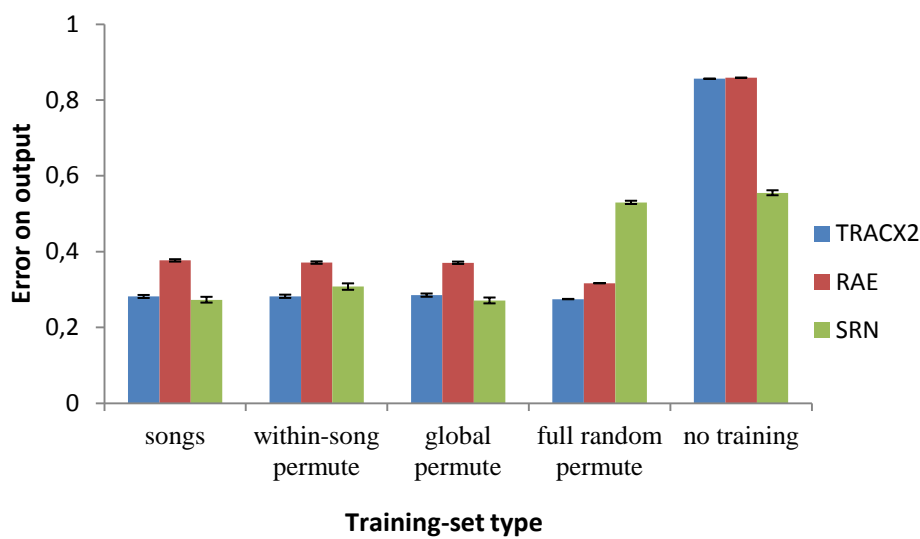

Figure 8 The effect of prior learning for TRACX2, RAE and an SRN on the recognition of words found in the Bach test set but not in the training corpus (SEM error bars)

It is interesting to note that for all three types of networks tested, it is the set of intervals in the training set, *regardless of their order*, that accounts for the recognition advantage of the words in the Bach sonata. This result is in agreement with a study (Tillmann & Bigand, 2001) that demonstrated that the temporal order of chords in the context sequence did not affect the harmonic priming effect on the final target chord.

### 5.2. The effect of prior learning on three different kinds of test items
#### 5.2.1. Method

To further examine the effect of prior learning on previously unseen-word processing, and, notably, the potential effect of proximity sensitivity, we investigated the response of TRACX2 to different kinds of words that it had never encountered during training on the children's songs. For this study, the network was trained on the primary corpus of children's songs (as described in Study 1), and then tested on a different set of materials that shared similar structural features, but were new and had not been previously encountered by the network. This approach mimics a general methodological approach used in music perception research, that relies on new (i.e., previously unheard) experimental items to test listener's music perception (e.g., Deliège, 2001; Schellenberg et al., 2002; Marmel et al., 2010). Creating specific experimental material that respects the same musical features as real-world music allows for investigating listeners' music perception in a controlled way, whether it tests for interval and contour processing (e.g., Schellenberg et al., 2002), tonal function (e.g., Marmel et al., 2010) or specific musical patterns or prototype-like cues (e.g., Deliege, 2001). In Deliege, 2001's study, listeners were first exposed to a given musical material and then tested for different items that were either old, new or modified on different dimensions and to varying degrees, thereby providing evidence for listeners' memory storage and its influence on perception (see also work by Dowling et al., 1995, 2001, 2014 testing short-term memory). Here, we adapted a similar approach for TRACX2 -- namely, after a training phase on a set of children songs, the model was tested with three types of new words that did not belong to the set of children's songs on which it was trained and were intentionally constructed to test the proximity sensitivity of the network. Specifically, three types of new words were created, notably words that were:

i) far from all the words encountered during training,
ii) close only to existing, but unfamiliar words,

iii) close to very familiar words.

Each of these three types of unheard words should produce different error profiles -- namely, the first category of unheard words will produce the largest errors, the second type of unheard words will sound somewhat familiar to TRACX2 and will, therefore, produce smaller errors than the unheard words in the first category, and the third type, being close to familiar words, will produce the lowest errors. The details of precisely what is meant by these three categories of unheard words and the definition of far versus close are as follows.

We define the Chebyshev distance (Cheb) between two words as the largest distance, measured in semitones, between the corresponding intervals of the two words. Consider the new word *caf*, which does not occur in the children's song set. The closest word to *caf* in the children's corpus is *jim*. Between *c* (-10 semitones) and *j* (-3 semitones) there is a difference of 7 semitones; between *a* (-12) and *i* (-4) there are 8 semitones, and between *f* (-7) and *m* (0) there are 7 semitones. Consequently, the Chebyshev distance between *caf* and *jim*, is 8, which we write as Cheb(*caf, jim*) = 8.

*i) When the unheard words are far from all the words encountered during training*

If the Chebyshev distance between two 3-interval words was greater than 5, we considered them to be "far apart". We looked at TRACX2's errors over a set of 50 invented words that were far from all of the words in the primary training corpus. So, for example, TRACX2's error-on-output for *caf* was 0.45. Given that the errors for all of the 3-interval words in the ten children songs in the primary corpus varied from 0.16 to 0.39 with an average of 0.22, an error of 0.45 can be considered as rather large.

*ii) When the unheard words are only close to unfamiliar words*

The unheard word, *osf*, for example, is close to the word *orf* (Cheb(*osf, orf*) = 1), which exists in the training set. However, *orf* occurs only once in the training set and, as a result, has an error-on-output of 0.26. This explains why the error on output of the very similar, but unheard word *osf* is 0.26, which is slightly more than 1 SD (0.036) above the average error value of 0.22 for all words in the training corpus. In other words, *osf*, a new word, is very similar to *orf*, which exists in the training corpus but was not well learned because of its low frequency.

*iii) When the unheard words were close to familiar words in the children songs.*

Consider *llm*, a word that never occurs in the children's songs, but is at a Chebyshev distance of 1 from *lmm*, and *mlm* in the training set. These two words occur 2 and 4 times, respectively, in the children's song set and have errors, 0.17 and 0.16, respectively, that are well below the mean error for all existing words. As expected, the error on the new word, *llm*, is low, with a value of 0.18.

We randomly generated three sets of 50 unheard words, corresponding to the above three categories of unheard words:
- 50 unheard words situated at a distance greater than 5 from all the words existing in the children's songs;
- 50 unheard words situated at a distance of 1 from existing, unfamiliar words, i.e., words with an error that was greater than the mean error + 0.5 SD. In other words, an error greater than 0.24 for TRACX2.
- 50 unheard words situated at a distance of 1 from existing, familiar words, i.e., words with an error that was less than the mean error − 0.5 SD. This translated as an error less than 0.2 for TRACX2.

5.2.2. Results

The mean errors for these three categories of unheard words were respectively 0.30, 0.26, and 0.20. ($F(2, 147) = 101.9$, $p<0.001$, $\eta_p^2 = 0.58$). A Tukey post-hoc analysis showed that all pairs of means were significantly different from each other (for all pairs, $p<0.001$).

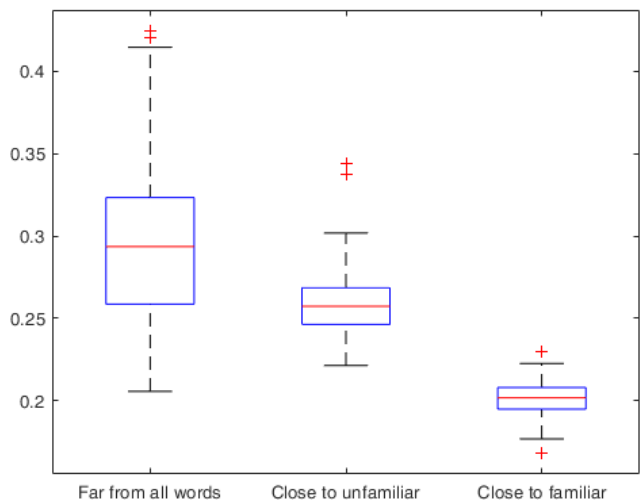

<u>Figure 9.</u> The effect of prior learning for TRACX2 on words of various distances from previously encountered familiar or unfamiliar words in the training corpus.

**5.3 Comparison with other models**

*First-order Markov models*

In this framework, TPs can only be estimated based on the observed frequencies of words present in the training set. For this reason, no generalization to new 2-interval words is possible. There is no straightforward means of estimating the corresponding transitional probabilities or of making use of a proxy, as is done by TRACX2, based on the proximity between intervals, a property that is not part of a simple first-order Markov model using TPs. The use of more sophisticated Markov models (e.g., dynamic n-order Markov models, Cornelius et al., 2017) is, however, beyond the scope of this paper.

*PARSER*

Perruchet (personal communication) tested PARSER (Perruchet & Vinter, 1998, 2002) by training it first on the primary set of children's songs and then testing it on the Bach sonata. He found no effect of prior learning on PARSER's chunk-extraction performance on the Bach sonata. Because PARSER is not equipped to handle distributed representations on input, it has no way of applying what it has learned about one 3-interval word in the training set to a similar, but never encountered word that appears in the test set. This is why there is no advantage of having been exposed to the children's songs prior to being tested on words in the Bach sonata.

*RAE*

An RAE shows a prior-learning effect for unheard words that is very similar to the effect for TRACX2. We tested this effect using the same paradigm we used for TRACX2 in 5.2.. The RAE was first trained on the primary set of children's songs for 30 epochs. We created three different sets of unheard words using the same procedure described in 5.2.1. We tested these three categories of unheard words with the RAE to determine its error-on-output.

The mean errors for the three categories of unheard words were respectively 0.46, 0.42, 0.33 ($F(2, 147) = 332$, $p<0.001$, $\eta_p^2 = 0.82$). The RAE, therefore, shows a similar prior-learning effect as TRACX2.

*SRN*

An SRN shows also a prior-learning effect for unheard words that is very similar to the effect for TRACX2. The SRN was tested in the same way as TRACX2 and RAE. The mean errors for the three categories of unheard words were respectively 0.39, 0.28, and 0.13 ($F(2, 147) = 69.2$; $p<0.001$, $\eta_p^2 = 0.49$).

**Conclusion**

In conclusion, TRACX2, RAE and SRN showed a significant effect of prior learning on the processing of new items that differed to various degrees from the items found in the training set. The finding that both the first-order Markov model and PARSER could not simulate these differences suggests the necessity of distributed representations to encode input. Further research will need to design new music material to be tested in perception experiments along this line in which errors-on-output of TRACX2, RAE and SRN will be used to predict listeners' performance in various recognition tests (e.g., lower errors predicting stronger confusion and thus lower accuracy). A similar approach has been previously used for the simulation of short-term memory results with the tonal-structure network being able to simulate participants' performance differences between the standard melody and four experimental conditions (i.e., exact transposition, tonal answer, atonal contour foil, random foil (Tillmann et al., 2000)). The outcome of our simulations here could be tested with an implicit learning-type experimental paradigm, notably an exposure phase followed by a test phase with targets and different foil types, applied to tone sequence material differing in interval use (similarly to the implicit learning experiment on 12-tone-music reported in Bigand & Poulin-Charronnat, 2006).

## 6. Study 3: TRACX2's sensitivity to melodic contours

As previously reported in music cognition research, human listeners are not only sensitive to the proximity of intervals (i.e., the distance between the corresponding intervals making up two sequences of intervals), but also to melodic contours (i.e., the "shapes" of the two sequences of intervals), even in infancy (e.g., Dowling, 1978; Trehub et al., 1985; Schellenberg, 1995). In this section, we will examine whether this sensitivity can be simulated with TRACX2.

### 6.1 Method

*Definition of a contour*

To address this question, we need to consider a rather subtle distinction, that of the *proximity* versus the *contour* of words. We have shown in §4.2. that TRACX2 is sensitive to the proximity of simple, two-interval words to the flat word *mm*. We have even argued, based on our grouping of the various 2-interval words, RR, R=, RF, =F, FF, F=, FR, and =R, and ==, that it might also be sensitive to contour information. In the following section we will tease apart the notions of proximity and contour and show that TRACX2 is sensitive, not only to proximity information, but to contour information as well.

To do this, we needed an operational definition of a contour. A contour can be simply defined as the sequence of rises and falls in a particular sequence of intervals. The contour of the word *kmo*, for instance, is ( - = +, which is read as Falling-Flat-Rising). For 3-interval words there are, therefore, 27 different possible contours.

One way to detect a contour effect would be to examine the internal representations of words of the same length. Those words belonging to the same contour should on average be closer together than those belonging to different contours. But we have already shown that TRACX2 is also sensitive to other factors, such as, the proximity of high-frequency words, and the location of the intervals inside a word (the trace of the final interval is stronger in the internal representation, see §4.3.). As a result, the study of contour effects can be biased by these other factors. To disentangle a potential contour effect from other effects, the pairs of words to be compared need to be carefully chosen.

Consider, for instance, the two words *sgm* and *okm*. They both have the same contour -- namely, (+ - =). They are composed of the intervals (6 -6 0) for *sgm* and (2 -2 0) for *okm*, which means that the Chebyshev distance (i.e., the largest distance between the two words across dimensions) between them is 4. However, the order of the intervals in the word matters, so we need to define a multidimensional distance, which we call **mdist**, between pairs of words. **mdist** is defined as the triplet of the absolute differences between the three intervals that compose the words. In other words, **mdist**(*sgm*, *okm*) = [4 4 0]. We now look at a 3-interval word whose **mdist** from *okm* is also [4 4 0] but that belongs to another contour, for instance, *kom* [-2,2,0] (Figure 10). If TRACX2 is, indeed, sensitive to contour information, we would expect the distance between the internal representations of *okm* and *sgm*, two words that belong to the same contour, to be smaller than the distance between the internal representations of *okm* and *kom*, that belong to different contours. This does, in fact, turn out to be the case. To show that this is true in general, we proceeded as follows:

- 1000 3-interval words were randomly generated. In order to keep these words "plausible", no interval above 12 or below -12 was considered and no sequence of two adjacent intervals with the same sign and adding to more than 12 or less than -12 were possible. This means there were no differences of consecutive notes going beyond one octave. For example, the following 3-interval words were not included: (0 13 5), (4 1 -13), (2 11 6), (5 -5 -8). But note that (11, 1, 11) would not have been rejected. This was done to keep the words "singable", or at least to avoid overly unusual melodic words.
- For each pair of words, we calculated the **mdist** between them and we noted the contour to which each word belonged.
- For a given mdist [a, b, c], all pairs of words with an mdist of [a,b,c] were selected.
- Among those pairs, some shared the same contour. These were put into a subset $S_1$. Pairs of words that did not share the same contour were put into a second set, S2.
- We then calculated the average cityblock distance between the representations of each pair belonging to S1. We did the same for each pair of words in S2.
- We compared the S1 distances to the S2 distances by means of an ANOVA.

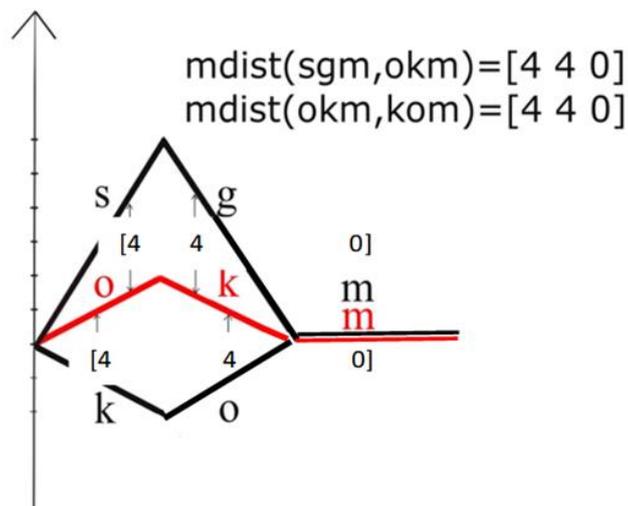

Figure 10. *sgm* and *kom* have the same **mdist** [4,4,0] from *okm,* but have different contours, (+,- =) and (-,+, =), respectively.

**6.2 Results**

In the 10 children songs in the primary familiarization corpus, the maximum cityblock distance between the internal representations of two 3-interval words is 44.4 and the average distance is 17.5. If we restrict ourselves to the pairs of 3-interval words that share the same contour, the average distance drops to 9.2. This average is computed on 165 pairs of words. (This was confirmed with the second set of children's songs where the distance dropped from 15.2 to 7.6.) This decrease would seem to reveal a contour effect. But the effect is not entirely convincing until the interval proximity between the words has been fully controlled for, as explained above.

The simulation with 1000 randomly generated 3-words made it possible to entirely eliminate the proximity effect. For an **mdist** of [2, 2, 2], for instance, we found 187 pairs of words with the same contour and 146 pairs with different contours. For the pairs of words belonging to the same contour, the average cityblock distance between their internal representations was 6.6, compared to 7.7 for the other 146 pairs. This difference is highly significant ($p < 0.001$), as revealed by an ANOVA.

We obtained similar results with other **mdist** values. We took all the triplets of mdist from [0, 0, 0] to [6, 6, 6]. This gave the expected result for 98% of the triplets. An ANOVA showed that differences were significant ($p<0.05$ with a Bonferroni correction) for 79% of all cases. Those results were confirmed on the second set of children songs (Set 2) where differences of all the triplets were in the expected direction (99%), and 96% of them were significant ($p<0.05$ with Bonferroni correction). As expected, without training there was no contour effect (2% of significant differences with $p<0.05$ with Bonferroni correction).

**6.3 Comparison with other models**

*First-order Markov-chain models.*

To the best of our knowledge, there is no explanation of the contour effect using first-order Markov-chain models. These models have no internal representations of the data they are processing and, as a result, no comparison is possible with the above results for TRACX2.

*PARSER*

PARSER does not construct internal representations of the data that are processed and no comparison is, therefore, possible with the above results for TRACX2.

*RAE*

With RAE we ran a simulation similar to the one carried out with TRACX2. 1000 randomly 3-words were randomly generated, and all the different triplets of **mdist**, from [0, 0, 0] to [6, 6, 6], were considered. This gave the expected result for 97% of the triplets. An ANOVA showed that differences were significant ($p<0.05$ with a Bonferroni correction) for 91 % of all cases. In other words, RAE was as contour sensitive as TRACX2.

*SRN*

An SRN also produces hidden-unit representations of the words in the training set, but the representations that it produces are considerably different from those produced by TRACX2, as explained in section §4.5.4. We ran the present contour-proximity simulation with the SRN and did not observe a contour effect. The differences were significant ($p<0.05$ with a Bonferroni correction) in the expected direction for less than 1% of all the cases. This result confirms the one reported in §4.5.4. where we already observed that the SRN clusters were far from the relatively disjoint clusters produced by TRACX2.

# 7. Study 4: Better recognition of the end of motives

Saffran et al. (1999) showed that participants are better able to recognize the end of melodic words than their beginning. Their results replicate a similar finding with speech stimuli (Saffran et al., 1996b) and suggest that the ends of words are learned first, whether the words are created from syllables or tones. Saffran and collaborators concluded, based on their results, that the transitional-probability learning mechanism that was posited to drive syllable-stream segmentation in infants (e.g., Saffran et al., 1996a; Aslin et al., 1998) could be the same learning mechanism as the one underlying tonal domains.

In this work, they began by defining a set of four tri-syllabic words (*abc*, *def*, *ghi*, *jkl*) made up of 12 distinct syllables (*a, b, c, d, e, f, g, h, i, j, k, l*). They then randomly concatenated these words with no immediate repetitions into a 2-minute familiarization sequence of 360 words. By means of a head-turn preference test, they compared infants' recognition performance to the original words versus "part-words", defined as the final syllable of one word followed by the first two syllables of another word. In general, however, the distinction between words and part-words in melody perception is not germane because sequences taken from real, pre-existing melodies do not consist of the concatenation of a pre-defined set of "tone-words" or "interval-words".  That said, in Saffran et al. (1999) tone-sequences were constructed, exactly mimicking the syllable-sequence construction in Saffran et al. (1996b). With respect to pre-existing melodies, they say, "The tone words were not constructed in accordance with the rules of standard musical composition and did not resemble any paradigmatic melodic fragments." After familiarization on this tone-sequence, infants were then tested for word/part-word discrimination as they had been in Saffran et al. (1996b).

We will now examine how well TRACX2 reproduces this asymmetry in the recognition of the "melodic" words used Saffran et al. (1999). This study is divided into three separate parts.

**7.1. Simulating the results of Saffran et al. (1999)**

### 7.1.1. Method

We began by attempting to reproduce the results observed in Experiment 3 of Saffran et al.'s (1999) human infant behavioral study. These authors constructed a tone sequence out of eleven pure tones in the same octave. The tones were combined into groups of three, thereby forming six "tone words." The tone words were: ADB, DFE. GG#A. FCF#, D#ED, and CC#D. The tone words were randomly concatenated with no immediate word repetition or acoustic markers, to create six different blocks, each containing 18 tone words. These blocks were then concatenated to produce a seven-minute continuous tone stream. There was no attempt to make tone words that resembled standard musical composition. In their analysis they define a "part-word" as being a three-tone sequence comprised of the two initial tones from one word plus a new third tone or the two final tones of a word plus a new initial tone.

All of our simulations were based, not on tone sequences, as in Saffran et al. (1999), but, rather, on interval sequences. Consequently, we replaced the 3-note words (and part-words) by 2-interval words (and part-words). Saffran et al. (1999, p.40) discussed at some length "the harmonic relations (intervals)" in their tone sequences. They showed the number of words containing particular intervals and how they differ. In other words, the authors were aware of potential confounds created by the overlapping intervals contained in their words.

As in Saffran et al., we created a training sequence by concatenating these 2-interval words. The problem we encountered, however, was that when we translated the L1 3-note words constructed by Saffran et al., into 2-interval words, this gave: *fv, un, hs, nl, nn,* and *pl*. And their 3-note part-words became our 2-interval part-words: *gv, pn, ls, nq, nw,* and *pn*. Clearly, the intervals *n* and *l* are overrepresented in these L1-words, with 4 repetitions for *n* and 2 repetitions for *l*. Further, *pn* was both an end-of-word and a beginning-of-word part-word. Saffran et al. (p. 41) writes "...we cannot rule out the possibility that interval information contributed to the tone segmentation process". Our simulations indeed confirm the importance of the interval information in the observed result patterns.

We, nonetheless, used these 2-interval words to produce a sequence as described in Saffran et al. and tested the errors produced when we tested the trained network on end-of-word (Xb) versus beginning-of-word (aX) part-words. The results of our simulations below suggest that interval information in their tone sequence may have indeed been a confound in the Saffran et al. experiments.

### 7.1.2. Results

As we pointed out above, *pn* can be a part-word that functions as either an end-of-word (Xb) or a beginning-of-word (aX) part-word. A first analysis considered it as an Xb part-word. After training TRACX2 for 100 epochs on the interval sequence created as described above, we considered the average of the errors-on-output of the three Xb part-words, {*gv, pn, ls*} and the two aX part-words, {*nq, nw*}. We averaged these errors over 20 runs of TRACX2 with a new interval sequence on each run. A paired-t test showed that the Xb errors were significantly smaller than the aX errors (t(19) = -2.38, p < 0.03, Cohen's d = -0.55, $BF_{10}$ > 2.2). In other words, TRACX2 reproduced the end-of-word advantage shown in Saffran et al. (1999) using a translation of the Saffran et al.'s 3-tone words into 2-interval words when *pn* is an Xb part-word.

However, because *pn* can be either an Xb or an aX, part-word, we removed it from the list of Xb part-words and made it an aX part-word. The new sets of part-words were, therefore, Xb = {*gv, ls*} and aX = {*pn, nq, nw*}. When this was done and we recalculated the average errors for the two types of partwords, the end-of-word advantage of Xb part-words over aX part-words disappears (p = 0.29). In other words, when *pn* was switched to an aX part-word, the significantly smaller errors of Xb part-words over aX part-words disappeared.

These seemingly contradictory results can reasonably be explained by the overabundance of the interval *n* in the training sequence. The fact that 25% of all intervals in the training set are *n* means that the error for any part-word containing *n* will necessarily be low. Thus, if *pn* is included in the Xb part-words, {*gv, pn, ls*}, its presence decreases the overall error for these part-words. Hence, the appearance of an end-of-word advantage. On the other hand, if *pn* is included among the aX part-words, {*nq, nw, pn*}, this significantly decreases the overall error of these part-words, thereby masking any potential end-of-word advantage of the Xb part-words.

In short, converting the sequence of 3-tone words used by Saffran et al. into an equivalent sequence of 2-interval words does not allow TRACX2 to systematically simulate their end-of-word part-word recognition advantage.

**7.2. Overcoming the problem of interval repetition**
7.2.1. Method

Because of the potential problem of interval repetitions in our interval encodings of Saffran et al.'s tone words, we created an interval-word sequence that satisfied the Saffran et al. sequence-creation methodology for tones, but did not have the interval-repetition problem described above. The 2-interval words with which we created the training sequence were: *fv, un, hs, dy, mt, pl*, and the associated 2-interval part-words on which we tested the network were: *gv, wn, rs, db, mo, pq*. We created a training sequence as in Saffran et al. (1999) and ran the program 20 times with 100 learning epochs, each time on a different training sequence constructed from the words. We compared errors on Xb part-words (i.e., {*gv, wn, rs*}) with those of the aX part-words (i.e., {*db, mo, pq*})[2].

7.2.2. Results

We averaged over the three Xb words and over the three aX words over 20 runs. A paired-t test showed that the Xb errors were significantly smaller than the aX errors (t(19) = -6.9, p < 0.001, Cohen's d = -1.55, $BF_{10}$ > 100). Saffran et al. reported that 64% of the time Xb part-words were recognized better than aX partwords. For TRACX2 in this case, this percentage was also 64%. In other words, with a sequence of intervals created with words that avoided the interval-repetition and *pn* part-word problem, TRACX2 reproduced the end-of-word advantage shown in Saffran et al. (1999). When trained on the above sequence, TRACX2 was, indeed, sensitive to the end-of-word advantage reported by Saffran et al. (1999). As it might be argued that this result might be overly dependent on the choice of the words making up the training sequence and the part-words, we turned to a third analysis based on TRACX2's internal representations to demonstrate and explain this advantage.

**7.3. Analyzing the end-of-word-advantage using the internal representations of the 3-interval words in the children songs**
7.3.1. Method

As the study of the internal representations built by TRACX2 revealed a similar bias towards the end of the words (see Section 2.2.2), we decided to address, in a third set of simulations, the end-of-word-advantage issue through the analysis of the internal representations of the 3-interval words in the children songs. For each 3-interval word – say *ayj* - we compared the internal representation of the full word (*ayi*) to the internal representations of its two first intervals, *ay*, and of its last two intervals, *yj*. The end-of-word preference revealed by Saffran et al. implies that the distance between the representation of the 2-interval word (*yj*) at the end of the full word and the representation of the full word, *ayj*, should be smaller than the corresponding distance between the representation of the 2-interval word, *ay*, at the beginning of the full word and the representation of the full word. In other words, Dist(*H(ayj)*, *H(yj)*) < Dist(*H(ayj)*, *H(ay)*), where *H* is the hidden-unit representation of the input vector of TRACX2 and Dist is the cityblock distance between two vectors. Even though other factors impact the way the internal representations are elaborated (frequency of occurrences, proximity, contours), the differences should emerge from the comparison of all the possible 3-interval words.

---

[2]Unlike Saffran et al., we did not create a second sequence made from the part-words of the first. We created a single sequence and tested the two types of part-words from that sequence.

### 7.3.2. Results

For each of the 161 3-interval words found in the children's songs, we calculated the cityblock distances between its internal representation and each of the two sub-words constituted by the first two and the last two intervals of the word. The average distance for the sub-word beginning the 3-interval words was 0.76 (0.77 for the second set of children songs) and for the sub-word ending the 3-interval word was 0.58 (0.60 for the second set of children songs). The effect was, in fact, observed on 90% of all 3-interval words (93% for the second set of songs). The direction of the mean difference was as announced by Saffran et al.'s observations.

### 7.4. Discussion

The sub-word asymmetry at the level of TRACX2's internal representations emerges naturally from the architecture of TRACX2, specifically from the fact that word accretion in TRACX2 involves adding individual items (whether they are syllables, images, or intervals) to the RHS of the input. Once again, consider the word *ayj*. The representation of the word is built in a hierarchical way. The two intervals, *ay,* are first chunked and the network's representation of the chunk, *H(ay)* is encapsulated in the LHS of the input. This means that the individual interval, *a*, making up *ay* has "disappeared" into the chunk *H(ay)*. Now, consider the sub-word, *yj*. When *y* and *j* are on input, its internal representation, *H(vj)*, will be closer to *H(ayj)* than *H(ay)* will be to *H(ayj)* because for both *ayj* and *yj,* the final interval, *j*, remains explicitly on the RHS of the input, whereas *ayj*'s initial interval, *a*, has been subsumed into *H(ay)*. This explains the smaller distance between *H(ayj)* and *H(yi)* compared to *H(ayj)* and *H(ay)*.

In other words, we do not need to invoke TPs or an anchor role played by the last note, as proposed by Saffran et al., to explain the end-of-word advantage effect. The chunk-accretion mechanism used by TRACX2 in which new items are added to the RHS of the input tends to better preserve the end of the chunks than their beginning, leading to the end-of-word advantage.

### 7.5. Comparison with other models

#### *First-order Markov-chain models*

Saffran et al.'s (1999) explanation of their results is in terms of TPs (of notes) which is the underlying mechanism of a first-order Markov-chain model explanation. In our simulation their explanation would require applying TPs to intervals rather than notes.

#### *PARSER*

When segmenting streams composed of pre-defined words as in Saffran et al. (1996a,b; 1999), PARSER, perhaps somewhat surprisingly, does not find part-words or, at least, only finds them extremely rarely (Perruchet, personal communication). For this reason, PARSER cannot be used to detect the end-of-word part-word advantage reported in Saffran et al. (1999).

*RAE*

We averaged over the three Xb words and over the three aX words over 20 runs using the sequence described in §7.2. A paired-t test showed that the Xb errors were significantly smaller than the aX errors (t(19) = -6.6, p < 0.001, Cohen's d = -1.48, $BF_{10} > 100$). Saffran et al. (1999) reported that 64% of the time Xb part-words were recognized better than aX partwords. This compared to 72% for RAE. In other words, when trained on the above sequence, RAE, like TRACX2, was, indeed, sensitive to the end-of-word advantage reported by Saffran et al. (1999).

The analysis of the RAE internal representations of the 3-interval words in the children songs made it also possible to reproduce the Saffran et al. end-of-word advantage found in §7.3. For each of the 161 3-interval words found in the songs, we calculated the cityblock distances between its internal representation and the two sub-words constituted by the first two and the last two intervals. The average distance for the sub-word beginning the 3-interval words was 0.53 (0.77 for the second set of children songs) and for the sub-word ending the 3-interval word was 0.37 (0.60 for the second set of children songs).

*SRN*

The SRN also reproduced the Saffran et al. end-of-word advantage when run on sequences constructed from the words, *fv, un, hs, dy, mt, pl*, and tested on the two sets of part-words, Xb = {*gv, wn, rs*}, and aX = {*db, mo, pq*} (see §7.2.). The effect with the SRN was far more pronounced than for TRACX2. Over 20 runs, Xb part-words were recognized better than aX part-words 80% of the time, compared to 64% for both Saffran et al. (1999) and for TRACX2. A paired-t test showed that the Xb errors were significantly smaller than the aX errors (t(19) = -68.1, p < 0.001, Cohen's d: -15.2, and a $BF_{10} > 100$).

We also ran for the SRN the simulation described in §7.3, in spite of the fact that the internal representation generated by the SRN are substantially different from those generated by TRACX2. The average distance for the sub-word beginning the 3-interval words was 0.11 (0.08 for the second set of children songs) and for the sub-word ending the 3-interval word was 0.04 (0.06 for the second set of children songs).

These simulations would also seem to support an end-of-word advantage.

# 8. General discussion

## 8.1. Overarching issues

The starting point for our work was TRACX (French et al., 2011) and TRACX2 (French & Cottrell, 2014; French & Mareschal, 2017; Mareschal & French, 2017), that have been used to successfully simulate a wide range of sequence segmentation and chunking phenomena from both the infant and adult literature on sequential verbal and visual materials. Our goal was to extend the use of this neural-network architecture in an attempt to capture segmentation and chunking of short melodic sequences.

Even if the model was initially designed to simulate syllable-based word perception (where there exists a clear distinction between words and non-words), it does not include a mechanism that makes a clear-cut difference between words and non-words. Indeed, the chunking mechanism modeled by both TRACX and TRACX2 makes it possible to build segments (referred to as "words") of different strengths (measured by their errors-on-output).

This made it appealing for simulation in a domain where a clear word/non-word distinction does not exist. Musical sequences, in particular melodies, are not built out of a pre-existing set of words out of which a melody is built. The boundary between previously-heard words and unheard words with very similar motives is decidedly blurry.

### 8.2 Summary of TRACX2's contributions to melody perception

The modeling of melody perception reported in this paper was carried out, not with the aim of developing a full model of music perception, but rather, to suggest that the type of mechanism implemented in the TRACX models -- namely, memory-based segmentation and chunking coupled with the re-utilization of the internal representations of the detected chunks -- may be a general cognitive mechanism underlying segmentation and chunking in vision, language, and music perception.

We have shown that phenomena observed in simple human melody perception and learning can be simulated by means of a recursive autoencoder neural network. It is crucial to note that our goal was not simply to devise an efficient algorithm or network to detect repeated sequences in a musical piece. That is best left to engineers. LSTM (Hochreiter & Schmidhuber, 1997) and other more sophisticated approaches, such as GPT-3 (Heaven, 2020), would clearly outperform TRACX2 in a music information retrieval task. Rather, our goal was to develop a cognitively plausible, emergent model of melodic sequence perception and melodic pattern acquisition. TRACX2 takes an unsupervised approach with no explicit rules or prior musical knowledge built into it (i.e., it does not incorporate information from music theory or empirical music perception data). Initially, all the connection weights in the network are small random numbers centered around 0. During learning, no external supervisor is used to train the connection weights and no explicit rules are applied. Segmentation and chunking emerge gradually. Internal representations of the input emerge from this bottom-up learning, and these representations then influence the perception of subsequent melodic sequences, thus simulating the cognitive top-down influences emerging from learned information. Previous simulations with TRACX2 have shown that these mechanisms can simulate human data for verbal and visual sequence learning, prediction and perception. Here we extend these simulations to musical material, thus providing converging evidence for TRACX2 as a cognitively plausible model that parsimoniously simulates data across modalities and materials.

The simulations presented here based on the mechanisms instantiated in TRACX2 provide insight into the way humans might detect and extract regularities from music and then use this acquired knowledge for perception, prediction and memory.

Among the phenomena that TRACX2 is able to simulate in a qualitatively accurate and psychologically plausible manner are:
- exposure to simple musical patterns on the ability to subsequently learn more complex patterns, even if these patterns have not been encountered previously ;
- the ability to learn a representation of melodic words that is sensitive to their contour;
- the higher sensitivity of the system to the ends of motives, which are better recognized and memorized than their beginnings.

The present simulations used the implementation of TRACX2, as reported in French & Cottrell (2014) and Mareschal & French (2017), with the only differences being (i) the type of input encoding used (ordinal encoding rather than one-hot encoding) and the use of intervals rather than notes, (ii) an error calculation that averages the errors for each of the consecutive pairs of intervals making up the word, and (iii) a modified ReLu squashing function, instead of the standard tanh function. For the work reported here we focus on relative pitch intervals as a simplifying assumption. Regarding melody perception, previous music cognition research has indeed shown that the perceptually relevant information is the relative pitch information and the emerging contour information, rather than the absolute pitch information (i.e., the encoding of the pitch of each individual element).

One of the key contributions of our paper is its demonstration of the necessity of "ordinal" encoding of the inputs instead of the one-hot encoding previously used by TRACX and TRACX2. Aside from the obvious problem of not encoding the amount of rising or falling of intervals (nor its size) with one-hot encoding, with ordinal encoding TRACX2's internal representations are richer in terms of the amount of information they store. When ordinal representations are used on input, the network's internal representations maintain a trace of the intervals making up words that it has encountered.

The simulations reported in Section 5 demonstrate the positive impact of early exposure to simple melodies on subsequent learning of more complex musical patterns. Our simulations showed that 2-interval words in a Bach sonata that did not appear anywhere in the training set of children's songs were, nonetheless, more easily perceived (i.e., had lower errors on output) when the network had been previously trained on children songs. This effect was also confirmed for another piece of "classical" music, a Chopin fantasy (simulations not reported). Additional simulations on words never heard by the system show the existence of an inheritance of familiarity by proximity that could explain the effect of exposure to melodies. The improved musical abilities of children with enhanced early exposure to music have been shown previously with music listening and musical activities (e.g., Hannon & Trainor, 2007; Gerry, Unrau & Trainor, 2012). This could also be seen as an example of network training that "starts small" (Elman, 1993) or of "incremental novelty exposure" during training (Alhama & Zuidema, 2018).

When examining TRACX2's internal representations after learning on a set of children's songs, we have also shown that the model is, indeed, sensitive to contour effects. To show this, we were able to factor out the influence of proximity, which is a confound in showing contour effects.

And finally, we have shown that TRACX2 simulates the end-of-word recognition advantage that was shown in Saffran et al. (1999). The conclusions drawn from these simulations were based both on error data from test sequences that we created according to the Saffran et al. word/part-word criteria, and, most importantly, the examination of the internal representations of the network.

Comparisons with other models showed that both first-order Markov chains and PARSER, which are both symbolic models, cannot reproduce all the results established with TRACX2. In particular, these two models do not generalize to unheard music. The SRN is substantially different from TRACX2 in both its architecture and its objective of predicting upcoming items in a sequence. We have shown that this leads to lower sensitivity to contours. This is arguably due to the fact that the SRN's prediction does not require explicit chunking of sub-sequences in the input stream. The comparison with RAE is more instructive. Indeed, TRACX2 and RAE differ only in how they chunk information. The chunking mechanism implemented in TRACX2 allows it to rapidly form distinct groups of its internal representations, which is not the case for RAE. Nonetheless, the performance of TRACX2 and RAE are similar, although not identical, on tasks involving familiarity judgments, priming effects, as well as end-of-word and contour effects. This is not surprising when only 2-interval words are considered. However, differences between the two systems appear on longer words where the chunking mechanism used by TRACX2 impacts the internal representations of those words (see §4.5.3). For these longer words, unheard-word familiarity is different for the two systems. Finally, the fit to real data with TRACX2 is better than for RAE, something that could be attributed to TRACX2's more sophisticated chunking mechanism. A more complete understanding of the differences between the two models will require additional studies. Interested readers are encouraged to contact the Corresponding Author to obtain the Matlab code for TRACX2 and the familiarization songs.

### 8.3. Limits of the model and future research

#### 8.3.1. Simplifications

As with any attempt to model a complex human ability, in this case, melody perception, there are limitations to what the TRACX2 model can do. Our simulations have only reproduced some well-known features of some simple elements of music perception. The levels of melodic-word familiarity, as measured by TRACX2's errors still need to be confirmed with new experimental data in future research. Further, the basic chunking mechanism of TRACX2 does not allow it to identify "singular motives", i.e., melodic words that are not repetitive, but, rather, stand out to human listeners because they are very different from what has been previously heard. This suggests that perhaps other basic (predictive) mechanisms, i.e., mechanisms more focused on the anticipation of what is coming – need to be integrated into TRACX2.

Our results were established on a simplified version of existing melodies. The next challenge for TRACX2 will be to use more complex musical information. In particular, information about the duration of the notes making up the intervals needs to be encoded in the input patterns. In our present simulations, half-notes, quarter-notes and eighth-notes, for instance, are not distinguished. Likewise, timbre, tonal-harmonic information (including chords), or even, pauses, were not part of the input encoding to TRACX2. One of the reasons that we felt that children's songs were an appropriate testbed for the model was because these songs can be recognized even without durational patterns (e.g., Devergie et al., 2010). Finally, the model does not take into account phenomena, such as, the role of attention, the musical culture of the listener, or memory-refresh mechanisms.

#### 8.3.2. Non-adjacent dependencies

TRACX2's chunking mechanism relies heavily on the sequential presentation of input data. Chunks are used only on the LHS of the input and, at least in the current instantiation of the model, the RHS can never contain a chunk, only an interval. This constrains the manner in which a chunk can be built: syllables, images, or intervals must be adjacent and chunks are formed by progressive accretion of single intervals and never already formed chunks identified in the input stream. Non-adjacent dependencies, where they might occur, are not chunked explicitly by TRACX2 in the same way that adjacent dependencies are. However, we have shown in §4.3 that, by means of a multiple correlational analysis of the network's internal representations, within long words non-adjacent dependencies are, indeed, captured by TRACX2.

Further, there are no attentional mechanisms in TRACX2 that would allow it to "focus" attention on certain intervals (e.g., *m*) or sequences of intervals, making them easier to remember or faster to learn, or to highlight non-adjacent dependencies (e.g., Creel et al., 2004).

#### 8.3.3. Future work

The TRACX2 model is, admittedly, just a starting point in the computational connectionist modeling of melody perception, but it provides a basis to generate new predictions for melody perception that then can be tested in targeted behavioral studies, including cross-cultural experiments. Experiments will need to be designed to compare melodic expectations with the results observed with TRACX2, to better understand the impact of the distribution of motives in songs, on how they are recognized, to assess the impact of proximity of contours on short-term memory, and to compare our results with those of other models of melodic perception and expectancy formation.

It is clear that purely bottom-up models will not be able to capture the full range of human music perception. Ultimately, modeling melody perception and adult music perception will necessarily involve an interaction between bottom-up learning (based on sensory input) and top-down control or predictions, such as, influences based on prior acquired knowledge, which can remain implicit, contain explicit rules and involve attention.

## 9. Conclusions

Our simulations suggest that the segmentation-and-chunking mechanism implemented in TRACX2 provides a plausible means of explaining some of the basic mechanisms of early music learning and perception. It combines a purely bottom-up approach with an emergent top-down mechanism -- namely, chunk-formation and the subsequent influence of these chunks on later perception. We believe that something like these learning and representational mechanisms could be used by a cognitive system to segment and chunk musical sequences during early music learning.

In addition, our present findings, taken together with previous research (French et al., 2011; Mareschal & French, 2017), suggest that the recursive autoencoder architecture implemented in TRACX2 could be a relatively domain-general mechanism, at least, insofar as it applies to domains beyond word segmentation and chunking (Frost et al., 2015). While the results presented in this paper have only scratched the surface of music perception, we believe that it is a first, fundamental step in the endeavor to understand the general mechanisms underlying human sequence processing.

To conclude, aside from the advantage of parsimony, the possibility of the existence of common mechanisms to explain linguistic, image and musical perception should not be underestimated. We believe that the underlying principles on which recursive autoencoders are based could lead to new predictions, new comparisons, better understanding and further insights into the mechanisms of perception and learning.

## 10. Acknowledgments

The authors would like to thank Bénédicte Poulin-Charronnat for useful discussions of harmonic priming. We would also like to thank Pierre Perruchet for running PARSER on our datasets, thereby allowing the performance of TRACX2 and PARSER to be compared. This research was supported in part by the Auditory Cognition and Psychoacoustics team of the Lyons Neuroscience Research Center which is part of the LabEx CeLyA ("Centre Lyonnais d'Acoustique", ANR-10-LABX-0060) at the University of Lyons.